\documentclass[11pt, a4paper, onecolumn]{article}
\usepackage[utf8]{inputenc}
\usepackage[margin=1in]{geometry}
\usepackage{subcaption}
\usepackage[numbers, sort&compress, super]{natbib}
\usepackage{graphicx}
\usepackage{booktabs}
\usepackage{amsmath}
\usepackage{hyperref}
\usepackage{xcolor}
\usepackage{lineno}
\usepackage{setspace}
\usepackage{float}
\newcommand{\sref}[1]{Supplementary Figure~\ref{#1}}

\newcommand{\MARCUSECGStanford}{87}
\newcommand{\MARCUSECGExternal}{91}
\newcommand{\MARCUSEchoStanford}{67.4}
\newcommand{\MARCUSEchoExternal}{86.0}
\newcommand{\MARCUSCMRStanford}{88.0}
\newcommand{\MARCUSCMRExternal}{85.0}
\newcommand{\MARCUSMultimodal}{70.0}
\newcommand{\TrainingImages}{13.5}
\newcommand{\SFTQuestions}{741{,}000}
\newcommand{\GRPOQuestions}{879{,}000}

\title{MARCUS: An Agentic, Multimodal Vision-Language Model for Cardiac Diagnosis and Management}

\author{
Jack W.\ O'Sullivan$^{1,2,*}$, Mohammad Asadi$^{3,*}$, Lennart Elbe$^{4,5,6}$, Akshay Chaudhari$^{7}$, \\
Tahoura Nedaee$^{8}$, Francois Haddad$^{1}$, Ivan Lopez$^{1,2}$, Fang Cao$^{1,2}$, Michael Salerno$^{4,5,6}$, Fei-Fei Li$^{9}$, \\
Ehsan Adeli$^{2,9,10}$, Rima Arnaout$^{4,5,6}$, Euan A.\ Ashley$^{1,2}$ \\[6pt]
{\small $^{1}$Division of Cardiology, Department of Medicine, Stanford University, CA, USA} \\
{\small $^{2}$Department of Biomedical Data Science, Stanford University, CA, USA} \\
{\small $^{3}$Department of Electrical Engineering, Stanford University, CA, USA} \\
{\small $^{4}$Department of Medicine, Radiology, and Pediatrics, UCSF, CA, USA} \\
{\small $^{5}$Bakar Institute, UCSF, CA, USA} \\
{\small $^{6}$UCSF--UC Berkeley Joint Program in Computational Precision Health, CA, USA} \\
{\small $^{7}$Department of Radiology, Stanford University, CA, USA} \\
{\small $^{8}$Department of Biology, Stanford University, CA, USA} \\[4pt]
{\small $^{9}$Department of Computer Science, Stanford University, CA, USA} \\
{\small $^{10}$Department of Psychiatry and Behavioral Sciences, Stanford University, CA, USA} \\
{\small $^{*}$Equal contributions}
}

\date{}

\begin{document}
\maketitle

\begin{abstract}
Cardiovascular disease remains the leading cause of global mortality. Progress is hindered by a critical bottleneck: human interpretation of complex cardiac tests. Artificial intelligence (AI) has emerged as a potential solution, however current vision-language models are limited to single-modality inputs, and are non-interactive. Here, we present MARCUS (Multimodal Autonomous Reasoning and Chat for Ultrasound and Signals), an agentic vision-language system designed for end-to-end interpretation of raw electrocardiograms (ECGs), echocardiograms, and cardiac magnetic resonance imaging (CMR) both independently and as multimodal input. MARCUS employs a hierarchical agentic architecture comprising modality-specific vision-language expert models, each integrating domain-trained visual encoders with multi-stage language model optimization, and coordinated by a multimodal agentic orchestrator. Trained on \TrainingImages{} million images (0.25 million ECGs, 1.3 million echocardiogram images, and 12 million CMR images), and our novel, expert-curated Q\&A and free-text dataset spanning 1.6 million questions, MARCUS achieves state-of-the-art performance surpassing frontier models (GPT-5 Thinking and Gemini 2.5 Pro Deep Think), and four single-modality cardiac AI models (ECG: EchoNext, Echocardiogram: EchoPrime and PanEcho, and CMR: ViTa). Across internal (Stanford) and external (UCSF) cohorts, MARCUS
achieves accuracies of 87--91\% for ECG, 67--86\% for echocardiography,
and 85--88\% for CMR interpretation, significantly outperforming
frontier models by 34--45 percentage points ($P<0.001$). Further, across five
public datasets spanning four countries, MARCUS also
outperforms four single-modality cardiac AI models by 6.3--35.2 percentage points ($P<0.001$). On complex multimodal cases, MARCUS achieves 70\% accuracy, almost triple the performance of frontier models (22--28\%) and produced interactive, free-text responses with 1.7--3.0$\times$ higher quality scores. Additionally, our agentic architecture confers resistance to mirage reasoning---the phenomenon whereby vision-language models derive reasoning traces from unintended textual signals or hallucinated visual content. MARCUS demonstrates that combining domain-specific visual encoders with an agentic orchestrator enables multimodal cardiac interpretation, offering a foundational tool for cardiologists worldwide. We release our trained models, code, and benchmark question test set open-source.
\end{abstract}

\section{Main}

Cardiovascular disease remains the leading cause of death worldwide, responsible for more than 20 million deaths annually\citep{dicesare2024}. While non-invasive diagnosis depends on the complementary diagnostic modalities of ECG, echocardiography, and CMR, the doubling of diagnostic volume over the past decade has outpaced clinical capacity\citep{papolos2016,goldfarb2021,tison2019}. With echocardiogram and CMR particularly data-rich (100 to 1000 images per study), expert human interpretation is time-intensive: 3--5 minutes for ECG\citep{kashou2023}, 20 minutes for echocardiogram\citep{wharton2015} and $>$30 minutes for CMR\citep{kramer2020}. Further, recent literature suggests that imaging contains (a) biomarkers of early diseases that routinely go unreported\citep{liu2023,blankemeier2022,zambrano2023} and (b) findings imperceptible to the human eye\citep{barry2023}. Compounding this increased clinical demand and time-intensive interpretation is the well-documented cardiology workforce shortage\citep{maddox2024}. With nearly half of all U.S.\ counties lacking a practicing cardiologist, timely interpretation of ECG, echocardiograms and CMR is expected to worsen\citep{kim2024}. Together, these factors underscore a pressing unmet need: scalable, timely, accurate and widely accessible tools that automatically, and near instantly assess multimodal cardiac data and assist clinicians to manage cardiac patients.

Artificial intelligence (AI) has emerged as a potential solution. Recent randomized controlled trials (RCTs) have established the utility of large language models (LLMs) in complex subspecialty care\citep{osullivan2026}. For instance, the Articulate Medical Intelligence Explorer (AMIE) upskilled generalists by reducing clinically significant errors from 24.3\% to 13.1\%\citep{osullivan2026}. However, even these vanguard systems remain ``text-only'' reasoning engines, fundamentally disconnected from raw physiological and imaging data. This limitation necessitates a manual human-text step and introduces the risk of upstream errors or omissions.

Subsequent advancements have sought to close this gap by developing models that interpret raw diagnostic signals directly\citep{christensen2024,ferreira2025,poterucha2025}. One example is EchoNext, a convolutional neural network model trained to predict probabilities of cardiac diseases from 12-lead ECG waveforms. EchoNext advances text-only approaches as it interprets the ECG directly in their raw, 12-lead form and achieved high accuracy in detecting cardiac disease. Similarly, EchoPrime is a video-based vision-language model that utilizes contrastive learning to perform comprehensive assessments of echocardiogram videos. EchoPrime can autonomously digest echocardiogram videos and generate reports by retrieving the most clinically relevant text from a corpus of prior interpretations\citep{vukadinovic2026}. Similarly, ViTa is a vision-transformer foundation model pretrained on
UK~Biobank cardiac MRI that predicts 18 quantitative cardiac phenotypes
directly from short- and long-axis cine sequences \citep{zhang2025}.

Despite these technical leaps, current vision models have a single-modality ceiling. EchoNext is architecturally restricted to ECGs, EchoPrime to echocardiograms, and ViTa to CMRs. This lack of cross-modal synthesis represents a critical limitation in clinical utility. These single modality tools help physicians with discrete tasks (`Based on this ECG does a patient need an echocardiogram'), but clinical cardiologists use multimodal data to diagnose, triage and manage a patient. To further advance clinical implementation, physicians and policy makers have recommended the next frontier of clinical AI models be multimodal\citep{osullivan2026,loriga2025,banerji2025,tu2025, miao2025}.

Further, in our companion work (Mirage)\citep{asadi2026mirage} we report the widespread phenomenon of ``mirage reasoning'' in current vision-language models. We demonstrate that mirage reasoning---where models derive reasoning traces from unintended text in preference to, and commonly without any reference at all to, provided images---is pervasive across current frontier models. For clinical application, future AI models must demonstrate not only multimodal synthesis but also immunity to mirage reasoning.

\begin{figure}[H]
\centering
\includegraphics[width=\textwidth]{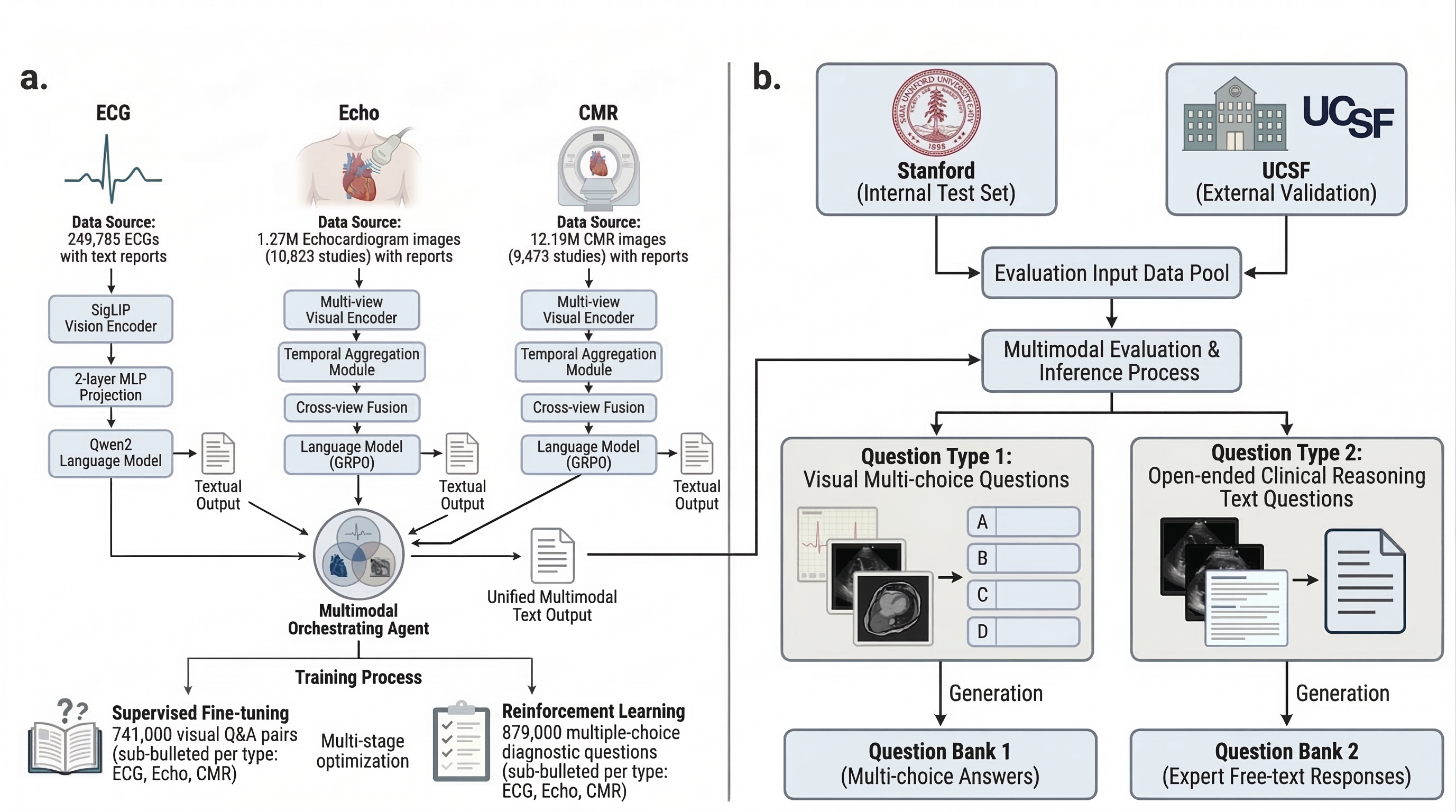}
\caption{\textbf{Study design and evaluation framework for MARCUS.} \textbf{a.}~Overview of the MARCUS architecture and training pipeline. The model integrates three cardiac diagnostic modalities: electrocardiography (ECG), echocardiography (Echo), and cardiac magnetic resonance (CMR). ECG data (0.25M ECGs) are processed via a SigLIP vision encoder and a 2-layer MLP projection into a 3B parameter language model. Echo (1.27M images) and CMR (12.19M images) data utilize multi-view visual encoders with temporal aggregation and cross-view fusion modules, feeding into a language model optimized via Group Relative Policy Optimization (GRPO). A Multimodal Orchestrating Agent synthesizes these inputs into a unified text output. The training process employs a multi-stage optimization strategy, including supervised fine-tuning on \SFTQuestions{} visual Q\&A pairs and GRPO reinforcement learning on \GRPOQuestions{} diagnostic multiple-choice questions. \textbf{b.}~Evaluation and validation workflow. The model's performance was benchmarked using an internal test set (Stanford) and an external validation cohort (UCSF). The evaluation encompasses two distinct tracks: (1) Visual Multi-choice Questions across all three cardiac modalities to assess precision, and (2) Open-ended Clinical Reasoning Text Questions to evaluate the model's capacity for expert-level free-text synthesis and interaction.}
\label{fig:overview}
\end{figure}

In this work, we present MARCUS (Multimodal Autonomous Reasoning and Chat for Ultrasound and Signals), an agentic vision-language system designed for automated interpretation and iterative reasoning across the three principal non-invasive cardiac modalities (ECG, echocardiogram and CMR). MARCUS employs a hierarchical architecture of modality-specific models, which use a vision transformer encoder and a 3-billion parameter vision-language model trained through a three-stage optimization pipeline comprising visual encoder pretraining on \TrainingImages{} million clinical images, supervised fine-tuning on \SFTQuestions{} expert-curated visual Q\&A pairs, and Group Relative Policy Optimization on \GRPOQuestions{} diagnostic questions. Modality-specific model outputs are then coordinated by an agentic orchestrator to achieve multimodal synthesis (Fig.~\ref{fig:overview}A). By fusing raw signals into visual tokens via multi-level cross-attention and residual connections, MARCUS achieves both multimodal synthesis and resistance to mirage reasoning. To rigorously evaluate MARCUS, we developed a novel benchmark of 1.6 million question-answer (Q\&A) and free-text reasoning tasks derived from 270,000 clinical studies. Validated against physician-verified reports, this framework quantifies both discrete diagnostic precision and high-level clinical synthesis, mirroring the iterative cognitive process of cardiologists.

MARCUS was trained on 249,785 ECGs; 1,266,144 echocardiogram images (10,823 studies), and 12,191,751 CMR images (9,473 studies) all with physician text ground truth reports. This foundation was refined through multi-stage optimization: supervised fine-tuning on a curated set of \SFTQuestions{} visual Q\&A pairs (460,000 ECG; 155,000 echocardiogram; 126,000 CMR) followed by reinforcement learning via \GRPOQuestions{} multiple-choice diagnostic questions (425,000 ECG; 236,000 echocardiogram; 218,000 CMR). We evaluated MARCUS across internal (Stanford), external (University of California, San Francisco (UCSF)), and five public datasets spanning four countries, demonstrating that MARCUS achieves state-of-the-art performance in single modality and multimodal accuracy, interactive reasoning (single and multimodal), generalizability and multimodal cardiac diagnosis. Furthermore, the agentic architecture confers resistance to mirages. Model code, weights (\url{https://github.com/AshleyLab/MARCUS}), and benchmark questions are available open-source.

\subsection{Expert single modality model accuracy exceeds frontier models}

\begin{figure}[H]
\centering
\includegraphics[width=\textwidth]{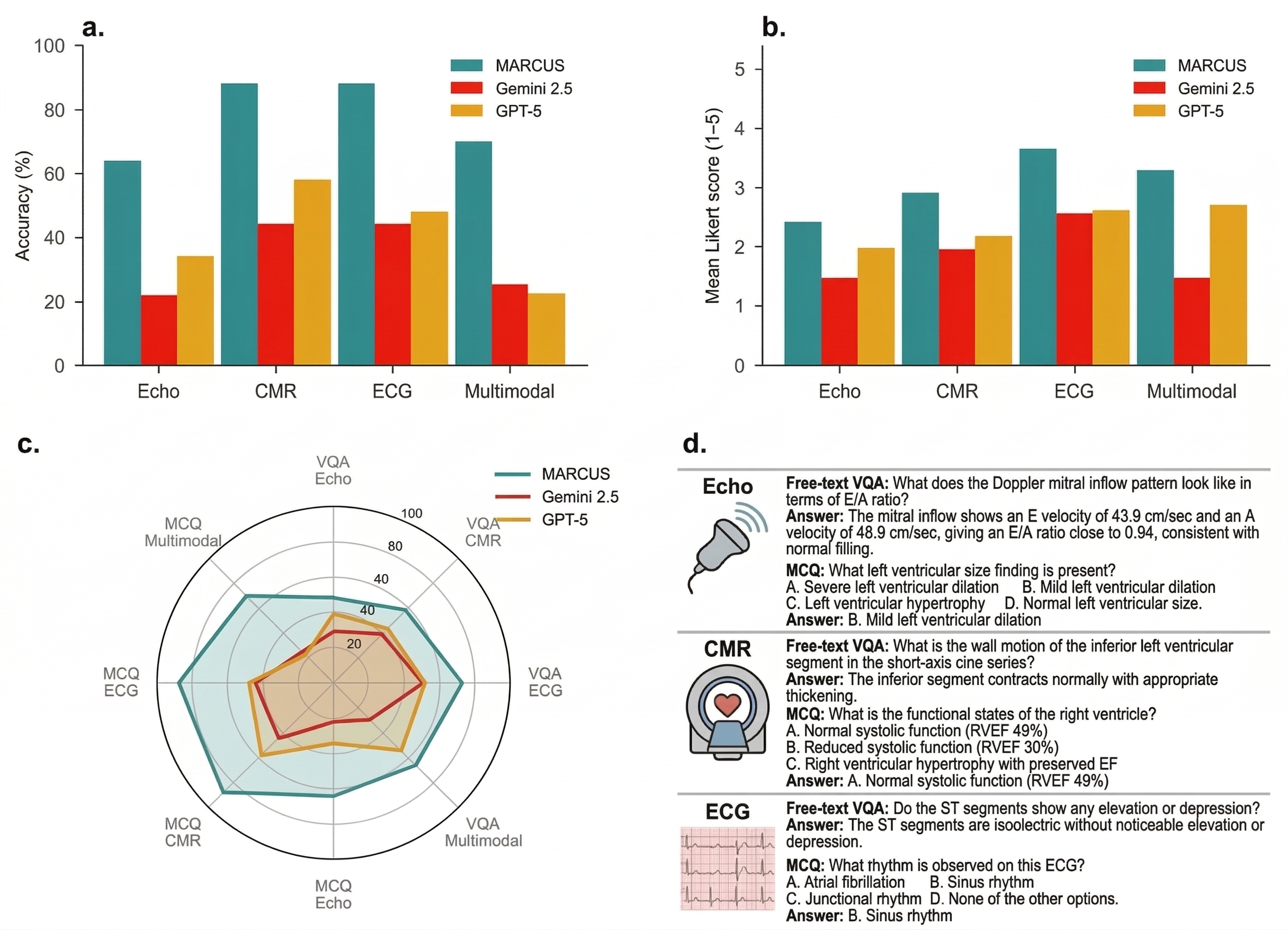}
\caption{\textbf{Comparative performance evaluation of MARCUS against frontier models.} \textbf{a.}~Accuracy (\%) of MARCUS, Gemini 2.5 (Deep Think), and GPT-5 (Thinking) on multiple-choice questions (MCQ) across four domains: Echocardiography (Echo), Cardiovascular Magnetic Resonance (CMR), Electrocardiography (ECG), and Multimodal (All). \textbf{b.}~Expert-rated performance on free-text clinical reasoning Visual Question Answering (VQA) tasks, represented by Mean Likert scores (scale 1--5). \textbf{c.}~Radar plot illustrating the holistic performance profile across all evaluation metrics. \textbf{d.}~Representative qualitative examples of VQA and MCQ prompts for Echo, CMR, and ECG, including model-generated answers.}
\label{fig:comparison}
\end{figure}

We first assessed the accuracy of MARCUS by evaluating the modality-specific expert models (ECG, echocardiogram and CMR) independently against two frontier models (GPT-5 Thinking and Gemini 2.5 Pro Deep Think). We determined the accuracy of MARCUS in answering modality specific visual multi-choice questions. We utilized a two-stage process to generate question-answer pairs: a cardiologist-curated Q\&A set was used as a template to prompt an LLM to create modality-specific question and answer sets with the corresponding ground truth answer extracted from the physician text reports. This same process was used to create a test visual multi-choice questions dataset at Stanford (internal test cohort, with none of these questions involved in training) and separately at UCSF (external test cohort), using each institution's respective data. Each visual multi-choice was accompanied by an image (ECG) or video (echocardiogram or CMR) with a corresponding question and 4--5 multiple choice answers, with one correct answer. All test set questions (at both Stanford and UCSF) were never seen before by any of the models. Across all three modalities, MARCUS demonstrated superior performance. For electrocardiograms (ECG), the model achieved an accuracy of \MARCUSECGStanford\% on the internal Stanford held-out test set and \MARCUSECGExternal\% on the external UCSF cohort, significantly outperforming the frontier models (frontier model accuracy: 35--48\% across Stanford and UCSF data; $P<0.001$ vs.\ MARCUS, Figure~\ref{fig:comparison}B). There was no secure HIPAA server to use Gemini at UCSF so external comparison was only MARCUS vs.\ GPT-5 (Thinking).

In echocardiography and cardiac magnetic resonance (CMR), the performance gap between MARCUS and frontier models widened. In echocardiography, which requires temporal reasoning, MARCUS achieved a diagnostic accuracy of \MARCUSEchoStanford\% in the Stanford cohort and \MARCUSEchoExternal\% in the external UCSF cohort (Figure~\ref{fig:comparison}B). This contrasted sharply with frontier models (accuracy 24--35\%, $P<0.001$ vs.\ MARCUS, Figure~\ref{fig:comparison}A and Figure~\ref{fig:comparison}B). Similarly, for CMR, MARCUS effectively digested and interpreted entire CMR studies including multi-slice cine and late gadolinium enhancement (LGE) imaging to achieve an accuracy of \MARCUSCMRStanford\% (Stanford) and \MARCUSCMRExternal\% (UCSF) (Figure~\ref{fig:comparison}B). Conversely, frontier models achieved an accuracy of 47--58\% ($P<0.001$ vs.\ MARCUS, Figure~\ref{fig:comparison}A and Figure~\ref{fig:comparison}B).

\subsection{Agentic orchestration enables synthesis of multimodal data}

While modality-specific models offer precision across ECG, echocardiogram and CMR individually, clinical cardiology relies on the synthesis of complementary data streams. To evaluate this capability, we tested MARCUS on a multimodal visual multiple choice question dataset that required the simultaneous integration of ECG, echocardiography, and CMR data to reach a correct conclusion. On this complex benchmark, MARCUS achieved a diagnostic accuracy of \MARCUSMultimodal\%, representing a fundamental performance leap over frontier models (Figure~\ref{fig:comparison}A). In contrast, GPT-5 (Thinking) and Gemini 2.5 Pro (Deep Think), when presented with the identical multimodal inputs and questions, achieved accuracies of 22.5\% and 27.5\%, respectively ($P<0.001$ vs MARCUS) (Figure~\ref{fig:comparison}A). This substantial performance gap (absolute difference $>$40 percentage points) highlights the agentic architecture of MARCUS, which decouples visual perception from clinical reasoning. Unlike standard models that attempt to map all inputs to a single latent space simultaneously, the MARCUS agentic orchestrator autonomously decomposes complex clinical queries into modality-specific sub-routines (e.g., ``Query ECG Expert: Is low voltage present?'' followed by ``Query Echo Expert: Estimate wall thickness'' and ``Query CMR expert: Is there late gadolinium enhancement''). By routing these sub-queries to the domain-specific expert encoders and aggregating their structured outputs, the orchestrator effectively mitigated the ``attention dilution'' often observed in frontier models. This approach allowed the system to preserve fine-grained visual details from specific modalities while constructing a coherent global diagnostic context, mirroring the clinical reasoning of a clinician. This capacity for synthesis was particularly evident in phenotypes where single-modality features are non-specific in isolation. A representative example is the distinction between constrictive pericarditis and restrictive cardiomyopathy. Echocardiography can often show biatrial enlargement, preserved ejection fraction, septal bounce, and respiratory variation in mitral inflow velocities. Confident diagnosis requires critical information provided by CMR (pericardial thickening, real-time septal shift during free breathing, lack of late gadolinium enhancement) and ECG (low voltage with nonspecific repolarization abnormalities). In contrast, frontier models frequently defaulted to the most salient single-modality feature, ignoring the contextualizing multimodal evidence. This ability to construct a unified patient representation from fragmented data streams validates the utility of agentic orchestration for higher-order clinical reasoning.

\subsection{MARCUS generates higher-quality, interactive clinical responses across single-modality and multimodal data}

Beyond discrete diagnostic classification, we evaluated the system's capacity to function as an interactive clinical consultant: answering open-ended and free-text queries. These open-ended questions were created in a similar way to the visual multiple choice questions. That is, an expert cardiologist developed a set of clinically relevant questions that were used as a template for an LLM to generate a large number of questions. Ground truth answers were extracted from physician text interpretations of respective ECGs/images/videos. Responses to these questions from MARCUS and frontier models were compared with ground truth physician text reports and results were reported via a mean Likert score. Likert score adjudication was completed by a blinded expert cardiologist (for a subset of questions) and also a separate LLM (that was not involved in question creation nor MARCUS training).

Across all single-modality tasks, MARCUS consistently generated responses of higher clinical value than frontier models. For open-ended ECG questions, MARCUS achieved a mean Likert score of 3.65 on the Stanford test cohort, compared to 2.60 for GPT-5 (Thinking) and 2.55 for Gemini 2.5 Pro (Deep Think) ($P<0.001$ vs MARCUS, Figure~\ref{fig:comparison}B). These results were replicated in the external UCSF cohort: mean Likert score for MARCUS: 4.14 and for GPT-5 (Thinking): 2.89, $P<0.001$ (Figure~\ref{fig:external}A). Similar performance margins were observed in the imaging modalities, with MARCUS outscoring frontier models in echocardiography (MARCUS: 2.41 vs.\ 1.97 GPT-5 Thinking and 1.47 Gemini 2.5 Pro (Deep Think), $P<0.001$ vs MARCUS, Figure~\ref{fig:comparison}B) and for CMR (MARCUS: 2.91 vs.\ 2.19 GPT-5 Thinking and 1.95 Gemini 2.5 Pro (Deep Think), $P<0.001$ vs MARCUS, Figure~\ref{fig:comparison}B). Results were replicated on the external UCSF cohort: Mean Likert scores for MARCUS of 3.24 for echocardiogram (vs 2.54 GPT-5 Thinking) and 3.22 for CMR (vs 2.60 GPT-5 Thinking) (Figure~\ref{fig:external}A).

In the multimodal setting, MARCUS achieved a composite quality score of 3.28 compared with 2.69 for GPT-5 Thinking and 1.46 for Gemini 2.5 Pro ($P<0.001$ vs MARCUS, Figure~\ref{fig:comparison}B). The performance difference likely reflected the agentic orchestrator within MARCUS, which successfully maintained context across modalities, generating reports that linked specific imaging findings to clinical recommendations (e.g., recommending the consideration of implantable cardioverter-defibrillator based on the combined evidence of LGE scarring on CMR and non-sustained ventricular tachycardia on ECG).

\begin{figure}[H]
\centering
\includegraphics[width=\textwidth]{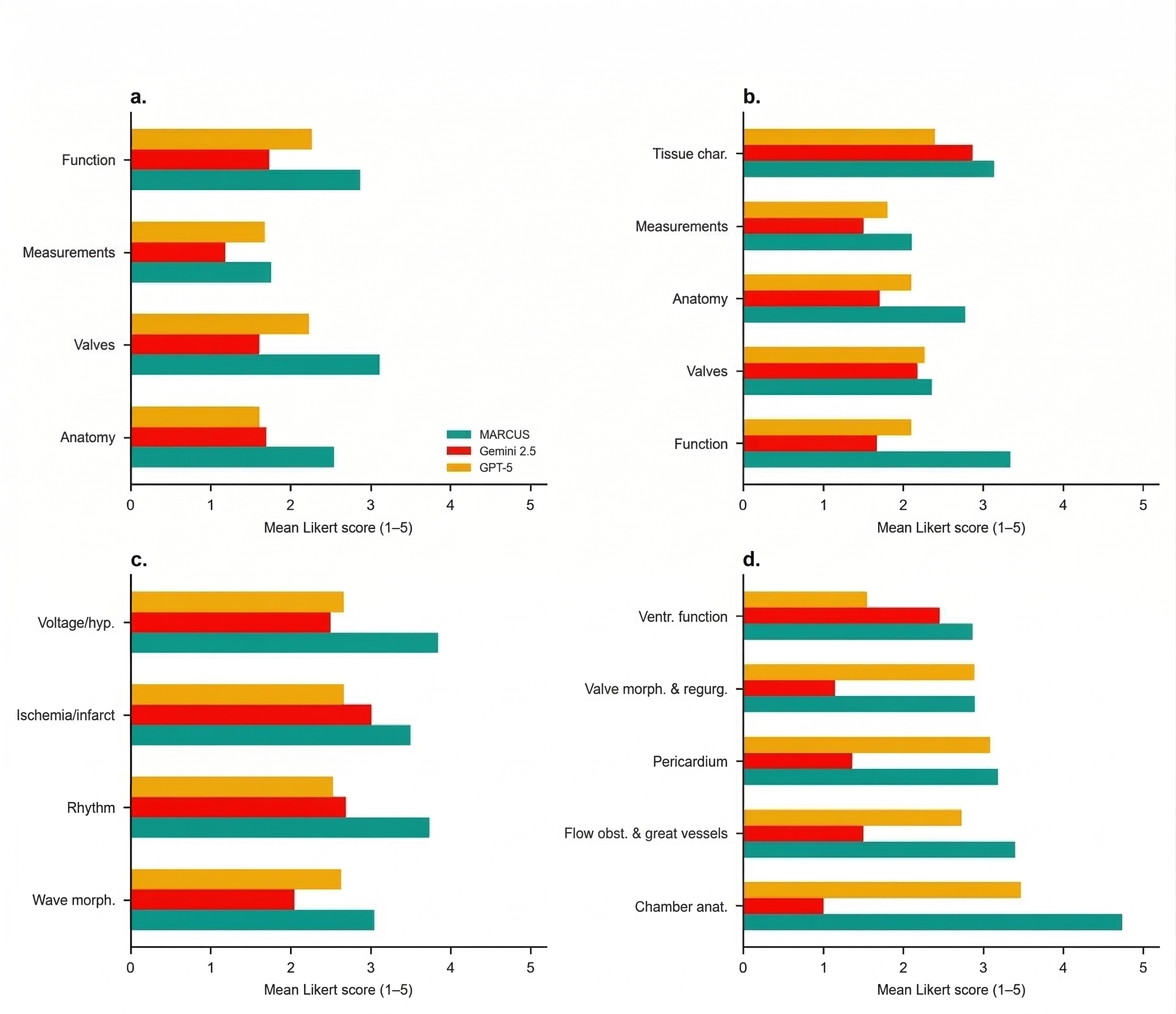}
\caption{\textbf{Granular performance breakdown across clinical sub-categories in free-text, open-ended clinical reasoning VQA.} \textbf{a--d.}~Mean Likert scores (scale 1--5) for MARCUS, Gemini 2.5, and GPT-5 evaluated across specific diagnostic domains within Echocardiography (a), Cardiovascular Magnetic Resonance (CMR) (b), Electrocardiography (ECG) (c), and Multimodal integration (d).}
\label{fig:subcategory}
\end{figure}

Subcategory analysis revealed that MARCUS demonstrated an advantage that was consistent across individual clinical domains within each imaging modality (Figure~\ref{fig:subcategory}). For echocardiography (Figure~\ref{fig:subcategory}A), MARCUS demonstrated its largest margin over frontier models in valvular assessment (MARCUS: 3.11 vs.\ GPT-5: 2.22 and Gemini 2.5 Pro: 1.61) and ventricular function (MARCUS: 2.87 vs.\ GPT-5: 2.27 and Gemini 2.5 Pro: 1.73). Performance was more closely matched across models for quantitative measurements, where all three models scored lower overall (MARCUS: 1.76, GPT-5: 1.68, Gemini 2.5 Pro: 1.19), suggesting that precise numerical retrieval from imaging data remains a shared limitation across current systems. For CMR (Figure~\ref{fig:subcategory}B), MARCUS performed best on pathology identification (MARCUS: 3.83 vs.\ GPT-5: 2.33 and Gemini 2.5 Pro: 2.17) and ventricular function assessment (MARCUS: 3.33 vs.\ GPT-5: 2.10 and Gemini 2.5 Pro: 1.67). Tissue characterisation was the most competitive CMR subcategory, with Gemini 2.5 Pro achieving a mean score of 2.87 compared to MARCUS 3.13 and GPT-5 2.40.

MARCUS demonstrated superior performance in ECG interpretation across all subcategories compared to frontier models (Figure~\ref{fig:subcategory}C). The largest absolute margins were observed in conduction abnormalities (MARCUS: 3.95 vs.\ GPT-5: 2.32 and Gemini 2.5 Pro: 2.53), rhythm interpretation (MARCUS: 3.74 vs.\ GPT-5: 2.53 and Gemini 2.5 Pro: 2.68), and voltage, hypertrophy and strain patterns (MARCUS: 3.83 vs.\ GPT-5: 2.67 and Gemini 2.5 Pro: 2.50). MARCUS also achieved near-perfect scores for arrhythmia identification (4.50) and ischaemia/infarction recognition (3.50), significantly outperforming GPT-5 (2.83 and 2.67, respectively) and Gemini 2.5 Pro (3.83 and 3.00).

\subsection{MARCUS maintains superior performance over frontier models across institutions}

\begin{figure}[H]
\centering
\includegraphics[width=\textwidth]{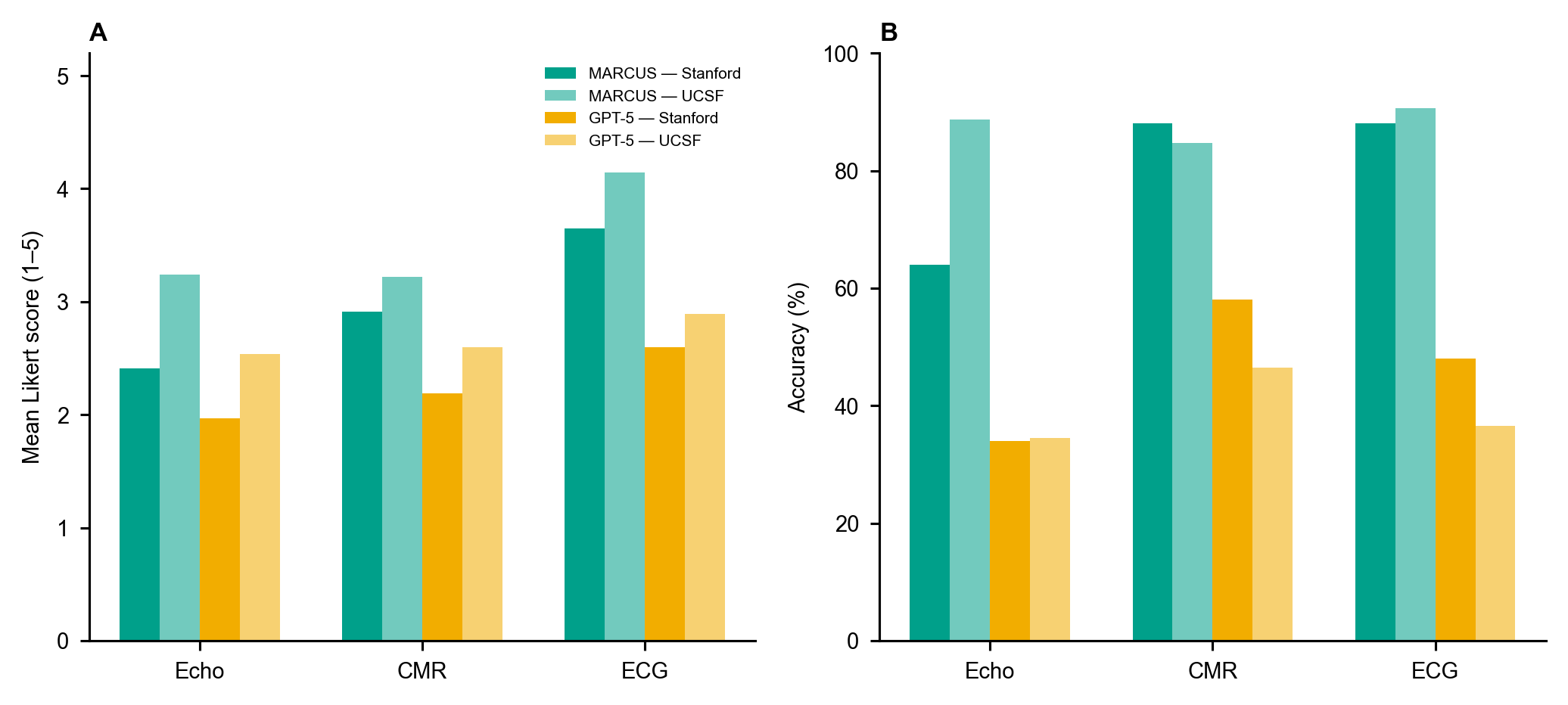}
\caption{\textbf{External validation of MARCUS and GPT-5 across Stanford and UCSF datasets.} \textbf{a.}~Visual question answering (VQA) performance, reported as mean Likert score (scale 1--5). \textbf{b.}~Multiple-choice question (MCQ) accuracy (\%). MARCUS substantially outperforms GPT-5 across all modalities and both task types, with consistent performance across institutions demonstrating robust external generalisability.}
\label{fig:external}
\end{figure}

A pervasive barrier to clinical AI deployment is performance degradation when algorithms are applied to new institutions. To assess robustness, we evaluated MARCUS on an external validation cohort from UCSF, distinct from the Stanford training site. There was no secure HIPAA server to use Gemini at UCSF so external comparison was only MARCUS vs.\ GPT-5 (Thinking). Despite differences in scanner vendors, imaging protocols, and patient demographics, MARCUS maintained diagnostic stability. For ECG and CMR, performance remained statistically equivalent across sites (ECG: \MARCUSECGStanford\% vs \MARCUSECGExternal\%, CMR: \MARCUSCMRStanford\% internal vs.\ \MARCUSCMRExternal\% external; $P>0.05$, Figure~\ref{fig:external}B). There was a difference in accuracy between the internal and external cohort for echocardiograms: \MARCUSEchoStanford\% vs \MARCUSEchoExternal\%, $P<0.05$, possibly pointing to the well-documented inter-institutional variability in image acquisition and interpretation\citep{morbach2018,felner1980}. However, the greater accuracy in the external cohort may suggest that MARCUS visual encoders learned invariant biological features rather than overfitting to site-specific artifacts. This robustness was unique to the agentic architecture; comparative frontier models plateaued at 35--52\% accuracy regardless of the testing site, MARCUS significantly outperformed these baselines in both environments ($P<0.001$), confirming that the system's ``clinical language'' generalizes across hospital systems.

\subsection{MARCUS outperforms frontier models in discrete diagnosis across all modalities}

\begin{figure}[H]
\centering
\includegraphics[width=\textwidth]{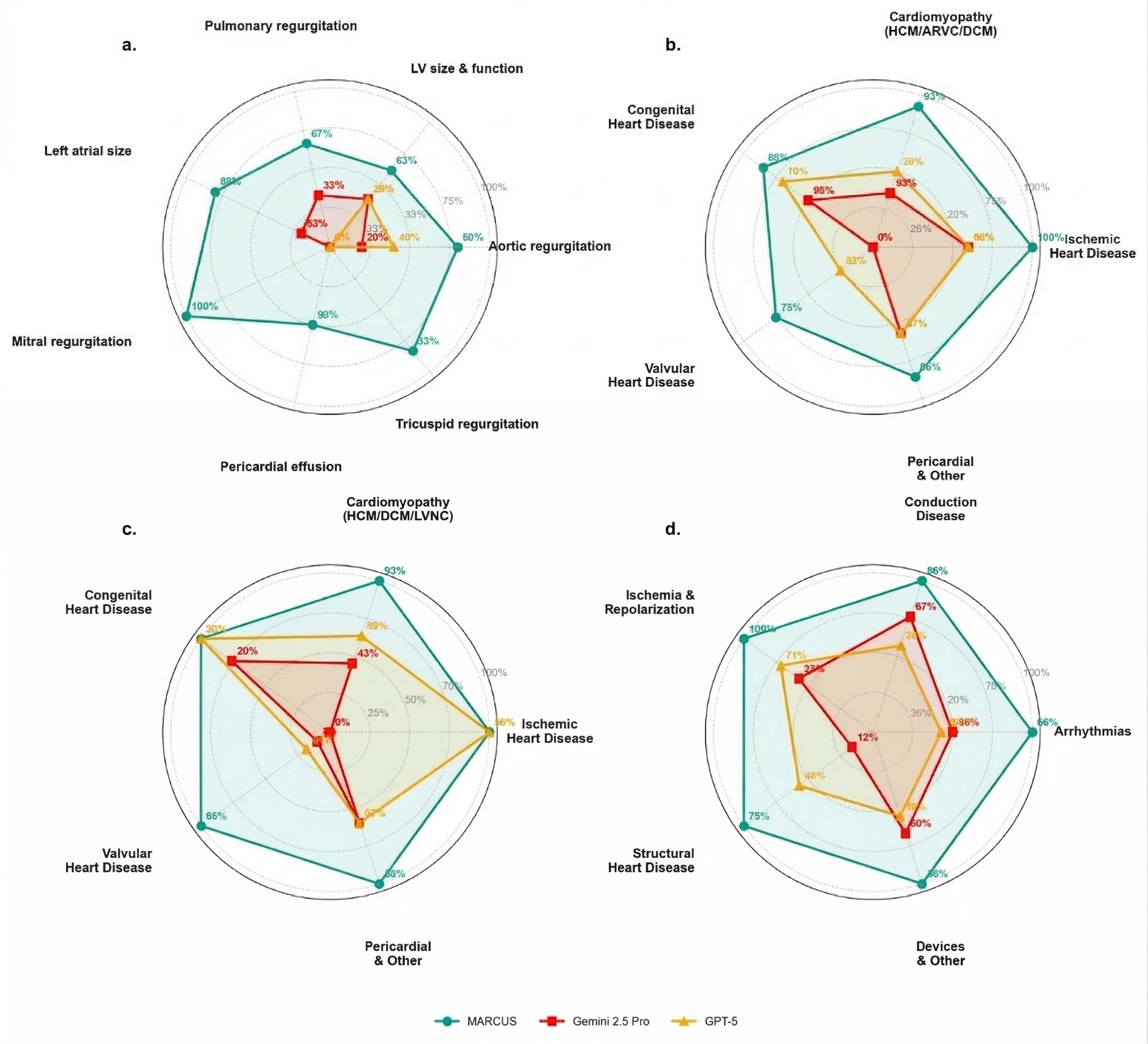}
\caption{\textbf{Discrete disease-category diagnostic accuracy for MARCUS, Gemini 2.5 Pro, and GPT-5 when answering multiple-choice questions (MCQs) on cardiac imaging.} \textbf{a.}~Multimodal. \textbf{b.}~CMR. \textbf{c.}~Echocardiography. \textbf{d.}~ECG. The radial gridlines indicate accuracy levels of 25\%, 50\%, 75\%, and 100\%.}
\label{fig:disease}
\end{figure}

A key task of AI systems is to reach discrete diagnoses. We tested if MARCUS had the ability to reach specific diagnoses across multimodal inputs. Across all modalities, MARCUS achieved superior diagnostic accuracy compared with Gemini 2.5 Pro and GPT-5 (Figure~\ref{fig:disease}). In multimodal cardiovascular assessment (Figure~\ref{fig:disease}A), MARCUS accuracy ranged from 50\% (pericardial effusion) to 100\% (mitral regurgitation). CMR performance was highest overall, with MARCUS achieving 100\% in ischemic heart disease and 85--93\% across cardiomyopathy, congenital, and pericardial categories, compared with 0--70\% for frontier models (Figure~\ref{fig:disease}B). Echocardiographic disease classification (Figure~\ref{fig:disease}C) demonstrated strong MARCUS performance in cardiomyopathy (92\%) and valvular heart disease (85\%), with lower accuracy in congenital heart disease and pericardial conditions (38\% each). ECG interpretation was consistently strong, with MARCUS achieving 88--100\% across all five categories versus 13--71\% for comparator models (Figure~\ref{fig:disease}D).

\subsection{MARCUS exceeds purpose-built specialist Cardiac AI models}
To test whether MARCUS's advantage over generalist frontier models extends to comparisons against purpose-built, single-modality AI systems, we benchmarked MARCUS against four task-specific models, each representing the current state-of-the-art (SOTA) for a single cardiac modality: EchoNext\citep{poterucha2025} (ECG), EchoPrime\citep{vukadinovic2026} and PanEcho\citep{holste2025} (Echocardiogram), and ViTa\citep{zhang2025} (CMR). For each single-modality model, we compared MARCUS across MCQ and free-text VQAs (reflecting clinical reasoning). We compared models across Stanford data, and five external public test sets spanning four countries: ACDC (Cardiac MRI; Dijon University Hospital, France)\citep{bernard2018}, the UK Biobank imaging substudy (Cardiac MRI; United Kingdom)\citep{sudlow2015}, MIMIC-IV-ECHO (Echocardiography; Beth Israel Deaconess Medical Centre, Boston, United States)\citep{johnson2023mimiciv}, MIMIC-IV-ECG (12-lead ECG; Beth Israel Deaconess Medical Centre, Boston, United States)\citep{gow2023ecg}, and PTB-XL (12-lead ECG; Physikalisch-Technische Bundesanstalt repository, Germany)\citep{wagner2020}. We computed accuracy with 95\% bootstrap confidence intervals and tested paired differences against MARCUS using McNemar's exact test on item-level responses from the Stanford test set.

\begin{figure}[H]
\centering
\includegraphics[width=\textwidth]{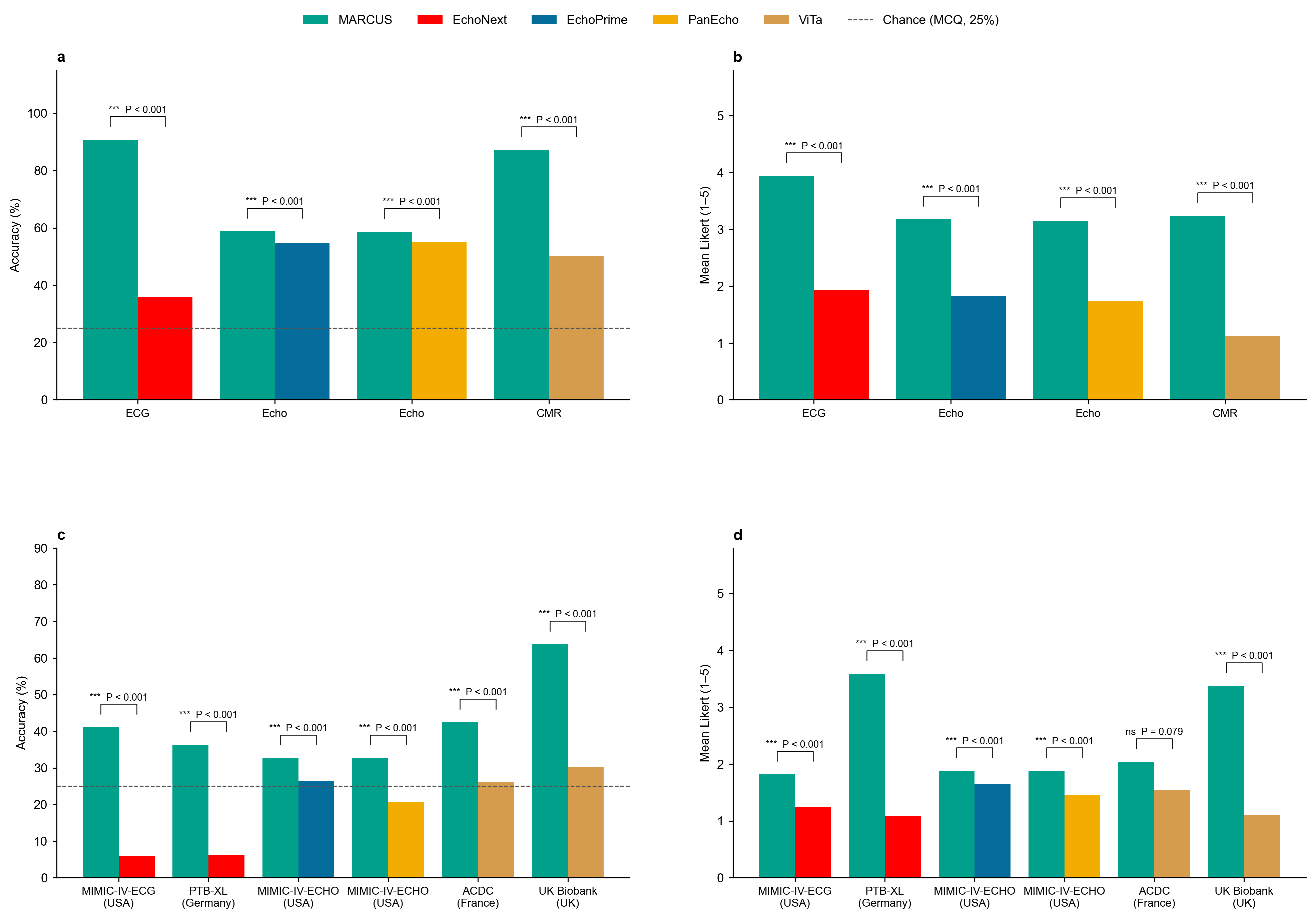}
\caption{\textbf{MARCUS versus purpose-built single-modality cardiac AI models, on the Stanford internal test set and on five external public test sets spanning four countries.} \textbf{a},~Multiple-choice accuracy (\%) on the Stanford test set for MARCUS vs each single modality model.
\textbf{b},~Free-text mean Likert (1--5; 1 = clinically wrong, 5 = fully correct) on the Stanford test set.
\textbf{c},~Multiple-choice accuracy on five external public test sets: MIMIC-IV-ECG and PTB-XL (12-lead ECG; USA and Germany), MIMIC-IV-ECHO (echocardiography; USA), and ACDC and UK~Biobank (cardiac MRI; France and the UK). \textbf{d},~Free-text mean Likert on the same five external test sets. EchoNext's twelve binary structural-heart-disease probabilities were formatted as a clinical pseudo-report (matching the procedure used for ViTa and PanEcho) and graded on the same 1--5 Likert rubric as MARCUS. Brackets indicate paired hypothesis tests with significance stars: McNemar's exact test for multiple choice (\textbf{a},\,\textbf{c}), Wilcoxon signed-rank test for mean Likert (\textbf{b},\,\textbf{d});
$^{***}P<0.001$, $^{**}P<0.01$, $^{*}P<0.05$, ns = not significant. Dashed grey line indicates chance performance on four-option multiple-choice questions.}
\label{fig:specialists_combined}
\end{figure}

For ECG interpretation, MARCUS substantially outperformed the SOTA ECG model (EchoNext\citep{poterucha2025}). Across multiple-choice questions on the held-out Stanford test set, MARCUS achieved 90.8\% accuracy compared with 35.9\% for EchoNext (absolute difference 54.9 percentage points, $P<0.001$) (Figure~\ref{fig:specialists_combined}a). The same pattern held on the two public ECG test sets: on the MIMIC-IV-ECG\citep{gow2023ecg} multiple-choice test set, MARCUS achieved 41.1\% accuracy versus 5.9\% for EchoNext (absolute difference 35.2 percentage points), and on the PTB-XL\citep{wagner2020} multiple-choice test set, MARCUS achieved 36.3\% versus 6.1\% (absolute difference 30.2 percentage points; (Figure~\ref{fig:specialists_combined}c). On free-text questions, MARCUS substantially outperformed EchoNext on every ECG test set. On the held-out Stanford free-text test set, MARCUS achieved a mean Likert of 3.94 (72.6\% Likert~$\geq 4$) versus 1.94 (4.8\% Likert~$\geq 4$) for EchoNext (Figure~\ref{fig:specialists_combined}b). On the held-out public MIMIC-IV-ECG free-text subset\citep{gow2023ecg}, MARCUS achieved a mean Likert of 1.82 (16.7\% Likert~$\geq 4$)
versus 1.25 (0.0\% Likert~$\geq 4$) for EchoNext. On the held-out PTB-XL free-text subset, MARCUS achieved a mean Likert of 3.59 (51.1\% Likert~$\geq 4$) versus 1.08 (0.0\% Likert~$\geq 4$) for EchoNext (Figure~\ref{fig:specialists_combined}d). 

For echocardiographic interpretation, MARCUS was benchmarked against two single-modality SOTA models, EchoPrime\citep{vukadinovic2026} and PanEcho\citep{holste2025}. On the held-out Stanford echo test set, MARCUS achieved 58.8\% MCQ accuracy versus 54.8\% for EchoPrime (absolute difference 4.0 percentage points, $P<0.001$) and 55.2\% (54.1--56.2) for PanEcho ($+3.5$~pp,$P<0.001$) (Figure~\ref{fig:specialists_combined}a). On the public MIMIC-IV-ECHO\citep{johnson2023mimiciv} test set, the same pattern held: MARCUS achieved 32.7\% MCQ accuracy versus 26.4\% for EchoPrime ($+6.3$~pp, $P<0.001$) and 20.8\% for PanEcho ($+11.9$~pp, $P<0.001$) (Figure~\ref{fig:specialists_combined}c). On free-text echocardiographic questions, MARCUS substantially outperformed both individual echocardiogram models. On the held-out Stanford free-text test set, MARCUS achieved a mean Likert of 3.18 versus 1.83 for EchoPrime ($P<0.001$) and 1.74 for PanEcho ($P<0.001$) (Figure~\ref{fig:specialists_combined}b). On the held-out MIMIC-IV-ECHO free-text questions\citep{johnson2023mimiciv}, MARCUS achieved a mean Likert of 1.88 versus 1.65 for EchoPrime ($P<0.001$) and 1.45 for PanEcho ($P<0.001$) (Figure~\ref{fig:specialists_combined}d). 

For CMR interpretation, MARCUS was compared against ViTa\citep{zhang2025}, a 3D+T vision-transformer foundation model pretrained on UK~Biobank cardiac MRI. On the held-out Stanford CMR test set, MARCUS achieved 87.2\% MCQ accuracy versus 50.0\% for ViTa (absolute difference 37.2 percentage points, $P<0.001$) (Figure~\ref{fig:specialists_combined}a). The same advantage held on two external, public CMR test sets. On ACDC\citep{bernard2018}, MARCUS achieved 42.5\% MCQ accuracy versus 26.0\% for ViTa ($+16.5$~pp, $P<0.001$) and on UK~Biobank\citep{zhang2025}, MARCUS achieved 63.8\% MCQ accuracy versus 30.3\% for ViTa ($+33.5$~pp, $P<0.001$) (Figure~\ref{fig:specialists_combined}c). On free-text CMR questions, MARCUS achieved a mean Likert of 3.24 on the held-out Stanford free-text test set versus 1.13 for ViTa ($P<0.001$) (Figure~\ref{fig:specialists_combined}b). The same advantage held on the two public CMR test sets: on ACDC, MARCUS achieved a mean Likert of 2.04 versus 1.55 for ViTa ($P<0.001$), and on UK~Biobank, MARCUS achieved a mean Likert of 3.38 versus 1.10 for ViTa ($P<0.001$) (Figure~\ref{fig:specialists_combined}d). 

Together these data demonstrate that MARCUS's agentic, multimodal architecture not only generalises beyond specialist coverage but also exceeds purpose-trained single-task models within those specialists' own domains. This performance extends beyond Stanford-source data into genuinely external multi-country test settings.

\subsection{Counterfactual probing by the agentic orchestrator confers mirage resistance for MARCUS}

In our companion paper\citep{asadi2026mirage}, we describe `mirage reasoning', the phenomenon in which vision-language models provide detailed descriptions of images never provided, including the generation of complex reasoning traces. We find that mirage reasoning is widespread amongst all vision-language models, fundamentally challenging current assumptions about visual understanding in these models. Here, we assess MARCUS's susceptibility to this effect across each modality-specific expert model independently using the counterfactual probing protocol described in Methods. When tested in isolation without the agentic orchestrator's verification, individual expert models exhibited non-zero mirage rates. When presented with image-absent queries, the ECG, echocardiography, and CMR expert models generated mirages in 33.0\%, 38.5\%, and 36.4\% of cases on average, respectively (\sref{fig:supp4}). These rates are consistent with the mirage effect reported across frontier VLMs\citep{asadi2026mirage}. However, when MARCUS was used to process multimodal data and thus had the full agentic pipeline engaged, the orchestrator's counterfactual probes successfully identified all occurrences of mirages. After confidence-weighted aggregation, MARCUS achieved a mirage rate of 0\% across all modalities and test cohorts, with no correctly grounded responses inadvertently suppressed. These results demonstrate that mirage susceptibility is not an intrinsic limitation of vision-language architectures and can be mitigated through inference-time verification, provided the system architecture supports counterfactual probing of its own components.

\section{Discussion}

In this work, we present MARCUS, an agentic, multimodal foundation model designed for the end-to-end interpretation of electrocardiograms (ECGs), echocardiograms, and cardiac magnetic resonance imaging (CMR). MARCUS represents an advance on current AI models in fundamental ways. First, MARCUS is capable of multimodal understanding, compared with existing, single modality models. Second, it is the first cardiovascular vision-language system capable of interactive clinical reasoning. Third, the multistage training pipeline employed in MARCUS (next-token prediction, supervised fine-tuning, and group relative policy optimization) coupled with its agentic orchestrator facilitates state-of-the-art performance in single and multimodality diagnostic accuracy, resistance to mirage reasoning, and produces novel, generative, non-contrastive output. Fourth, MARCUS is superior to frontier models across multimodal and single modality interpretation, and is superior to single modality cardiac AI models across six datasets spanning four countries. Fifth, with MARCUS we introduce a large, rigorous Q\&A benchmark curated by expert cardiologists and coupled with ground truths extracted from physician reports. Finally, MARCUS is trained on the largest corpus of multimodal cardiac data (\TrainingImages{} million images) to date. To accelerate research and clinical translation, we have publicly released our trained models, code, multimodal dataset, and visual questions benchmarking set as an open-source resource for the scientific community.

The substantial performance margin between MARCUS and frontier models, averaging 34--45 percentage points in single-modality tasks and widening to a 43--48 point gap in multimodal settings, highlights a critical ``clinical data gap'' in current AI development. While general-purpose frontier models are trained on internet-scale datasets, they remain shielded from the high-fidelity, longitudinal data stored within hospitals. This ``data illiteracy'' is most evident in the diversity of our results: while frontier models struggled with raw signal perception (achieving only 22--28\% accuracy on complex multimodal cases), the exposure to \TrainingImages{} million clinical images allowed MARCUS to reach 70\% accuracy. This advantage extends beyond discrete classification into higher-order clinical synthesis; the 1.7$\times$ to 5-fold increase in Likert scores for interactive reasoning suggests that the ``intelligence'' of a cardiovascular AI is as much a function of institutional data access as its architectural parameters. Ultimately, these findings underscore that the next frontier of medical AI must focus on learning directly from the raw, multi-layered, and multimodal clinical data.

Previous AI models in cardiology have excelled at discrete tasks using single modality inputs, such as detecting structural heart disease from ECGs\citep{attia2019}, automating echocardiographic measurements\citep{vukadinovic2024,holste2025}, or identifying cardiomyopathies from CMR\citep{ferreira2025,zhang2025,datar2024}. These models have used discriminative architectures that output fixed classifications or measurements or are CLIP-style contrastive models that retrieve historical report text. MARCUS advances these approaches by producing generative, novel and patient-specific output. By using the multistage training of next-token prediction, supervised fine-tuning, and GRPO, MARCUS produces novel natural language. This enables interactive clinical reasoning (which is not available in any prior model): clinicians can query the model, request explanations, and engage in iterative diagnostic dialogue. Importantly, MARCUS advances the cardiac AI field as the first dedicated multimodal model, but also MARCUS has higher accuracy and superior clinical reasoning than single-modality cardiac AI systems, which were designed exclusively to interpret their respective single modality (e.g. Echocardiograms for EchoPrime). MARCUS's superior performance was robust across seven datasets from six separate institutions across four countries. 

The mirage phenomenon described in our companion paper\citep{asadi2026mirage} poses a fundamental challenge for clinical deployment of vision-language models: if a model can generate plausible interpretations without referencing the image, clinicians have no reliable way to distinguish grounded reasoning from confabulation. Our results suggest that mirage susceptibility is not an intrinsic limitation of vision-language architectures but rather a consequence of end-to-end systems that lack intermediate checkpoints for verifying visual grounding. The agentic architecture within MARCUS provides such checkpoints.

Our work has limitations. Our training data was derived from a single centre. While we validated performance externally, generalizability to diverse community populations, imaging protocols, and equipment vendors requires further study. Second, our evaluation centered on curated multiple-choice and structured clinical reasoning open-ended questions. This format enables rigorous benchmarking, but may not fully capture the heterogeneity of real-world clinical queries or the ambiguity inherent in many imaging findings. Third, this study is retrospective. It remains unclear whether MARCUS improves outcomes, reduces diagnostic errors, or shortens time to treatment. These remain to be established in prospective trials. Notably, these limitations are not unique to our work; single-centre development, curated evaluation datasets, and retrospective design characterize most published AI models, underscoring the need for standardized, multi-institutional benchmarks and prospective validation frameworks across the field.

In conclusion, MARCUS represents a step towards AI systems that reason across modalities central to modern cardiovascular care. By moving beyond single-task classification to interactive, multimodal interpretation, this work offers a foundation for the next generation of clinical decision support; a paradigm where AI augments clinical care and extends specialist-level assessments to patients near-instantly and to those who might not otherwise have access.

\section{Methods}

\subsection{Overview}

We created interactive, vision-to-language models for each of our three cardiac modalities (ECGs, echocardiograms, and CMRs) separately and a multimodality orchestrating agent that synthesizes all three data types. The echocardiogram and CMR individual encoder models followed a similar architecture and comprised four main components: (1) a multi-view visual encoder that extracts imaging features, (2) a temporal aggregation module that models cardiac motion, (3) a cross-view fusion mechanism that integrates information across views, and (4) a language model that generates textual outputs with Group Relative Policy Optimization (GRPO) (Supplementary Figure~1). The ECG model consisted of three main components: (1) a SigLIP vision encoder that extracts features from patch-based image representations, (2) a two-layer MLP projection module that aligns visual embeddings with the language model's representational space, and (3) a Qwen2 language model that generates textual outputs. The study was approved by Stanford's Institutional Review Board (IRB: 67663, 41045).

\subsection{ECG, echocardiogram and CMR data processing}

ECG tracings were extracted from institutional XML files and converted to standardized image representations. Raw signals were reconstructed as images from extracted arrays, maintaining standard calibration parameters (25~mm/s, 10~mm/mV) as reference. To enable compatibility with vision transformer architectures, which process images at fixed resolutions ($224\times224$ pixels per patch), 12-lead ECG data were reformatted into a $4\times3$ grid layout. This transformation addressed competing constraints: standard ECG images display approximately 2.5 seconds per lead due to paper width limitations (250~mm at 25~mm/s), whereas comprehensive interpretation requires visualization of the full 10-second recording. To accommodate the complete recording within each lead's allocated pixel space while preserving interpretability, the time axis was compressed, altering the apparent mm/s ratio in pixel space. Amplitude was similarly scaled to ensure voltage deflections remained discernible following downsampling, modifying the effective mm/mV ratio. Spatial relationships between signal and background grid were preserved through proportional scaling. Although absolute calibration parameters do not directly translate to pixel space, relative calibration (the relationship between waveform amplitude and grid spacing) remains consistent across all processed images.

Echocardiographic studies were extracted as DICOM video files comprising multiple acquisition views per study. Unlike CMR, echocardiographic files lacked standardized metadata for automated view identification. To address this, we developed a view selection pipeline using visual attention. All videos from a single study were arranged into a composite grid, and an attention-based mechanism was applied to identify and weight relevant views for each clinical query. This approach enabled the model to dynamically prioritize diagnostically relevant acquisitions (e.g., apical four-chamber views for left ventricular function, parasternal long-axis views for aortic valve assessment) without requiring manual annotation or metadata-based filtering. Each selected video was then decomposed into constituent frames, with frames divided into $16\times16$ patches for processing by the vision encoder. Sequential feeding of frame-level tokens preserved temporal information through the language model's positional encoding, enabling capture of cardiac motion dynamics across the cardiac cycle.

CMR studies were extracted as DICOM files with accompanying metadata specifying acquisition parameters, sequence type, and anatomical plane. Unlike echocardiography, CMR metadata enabled automated sequence selection: the language model component parsed available metadata and selected appropriate sequences based on the clinical query (e.g., cine sequences for functional assessment, late gadolinium enhancement for fibrosis characterization, short-axis stacks for volumetric quantification). This metadata-driven routing allowed the model to autonomously identify relevant imaging planes without requiring exhaustive processing of all acquired sequences. Selected sequences were then decomposed into individual slices and frames, with each frame divided into $16\times16$ patches for vision encoder processing. For multi-slice acquisitions (e.g., short-axis stacks), spatial relationships across slices were preserved through sequential token ordering, enabling the model to reconstruct three-dimensional cardiac geometry. Temporal dynamics within cine sequences were similarly encoded through positional embeddings of frame-ordered tokens.

\subsection{Question and answer datasets}

We curated large-scale question-answer datasets for each cardiac modality from a single academic medical center (Stanford University). Source data comprised complete imaging studies: 249,785 ECGs; 1,266,144 echocardiogram images (10,823 echocardiogram studies), 12,191,751 CMR images (9,473 CMR studies) with paired physician reports and linked electronic health record diagnoses.

Question-answer pairs were generated through a two-stage process. First, an expert cardiologist developed a template library of 100 clinically relevant question types spanning diagnostic findings, quantitative measurements, and clinical recommendations. Second, physician text reports were processed by a large language model prompted to generate modality-specific questions, with ground-truth answers extracted directly from the corresponding reports. For multiple-choice questions, an expert cardiologist reviewed each item and created plausible distractor options (e.g.\ ``What is the left ventricular ejection fraction?'', correct answer ``mildly reduced'' with distractors ``normal,'' ``moderately reduced,'' ``severely reduced'').

This process yielded \SFTQuestions{} visual question-answer pairs for supervised fine-tuning (460,000 ECG; 155,000 echocardiogram; 126,000 CMR) and \GRPOQuestions{} multiple-choice diagnostic questions for GRPO reinforcement learning optimization (425,000 ECG; 236,000 echocardiogram; 218,000 CMR). For external validation, an independent dataset was curated from UCSF using identical question-generation methodology applied to locally acquired imaging studies and physician reports.

Open-ended questions were designed to assess clinical reasoning beyond discrete classification, requiring models to synthesize findings, generate differential diagnoses, and propose management recommendations. Ground-truth responses were derived from the assessment and plan sections of physician reports.

\subsection{Model training overview}

MARCUS is a generative agentic vision-language model that autonomously interprets ECGs, full echocardiogram studies (video form), and full CMR studies (also video form) as single-, dual-, or three-modality inputs. To build MARCUS, we undertook three steps: (1) Pre-training of modality-specific vision encoders on the full imaging corpus with paired physician text reports as ground truths; (2) Question and answer specific fine-tuning on our novel Q\&A datasets; (3) Group-Relative Policy Optimization (GRPO) to further enhance model precision by optimizing reasoning traces that led to correct final diagnoses (Figure~\ref{fig:overview}A).

We constructed MARCUS to be capable of interpreting three modalities individually (ECG, echocardiogram, and CMR) and collectively (multimodal). To achieve multimodal capacity we created an agentic orchestrator that incorporates all single modality outputs, and combines the text output into an overall assessment of the patient (Figure~\ref{fig:overview}A, Supplementary Figure~2).

We evaluated MARCUS on both single-modality and multi-modality tasks. We compared MARCUS's performance answering multiple choice and open-ended questions to that of large, frontier models: GPT-5 (Thinking)\citep{singh2025}, and Gemini 2.5 Pro (Deep Think)\citep{comanici2025}.

\subsection{Model architecture and training: Single modality ECG model}

A vision-language model for ECG interpretation was developed using a LLaVA-based architecture comprising a SigLIP vision encoder, two-layer MLP projection module, and Qwen2 language model. ECG signals were extracted from XML files, reconstructed as images, and reformatted into a $4\times3$ grid displaying the full 10-second, 12-lead acquisition. Grid lines were scaled proportionally with the signal to preserve relative calibration between waveform deflections and background spacing. Each image was divided into patches processed independently by the vision encoder, with the projection layer transforming patch embeddings into language model tokens (Supplementary Figure~3).

Training followed a two-stage procedure. First, the language model weights were frozen while the vision encoder and projection layer were trained, establishing a modality-specific visual encoder for ECG representation. Subsequently, both vision and language components were jointly fine-tuned on the complete dataset to produce the vision-language model. The final model was further fine-tuned using Group Relative Policy Optimization (GRPO), a reinforcement learning technique prioritizing correct answers/final diagnosis. Models were trained on interleaved $\langle$question$\rangle$$\langle$image$\rangle$$\langle$answer$\rangle$ sequences pairing clinical queries with ECG images and expert-derived responses.

\subsection{Model architecture and training: Single modality echocardiography and CMR model}

The echocardiography and CMR models followed a similar architectural framework but incorporated additional components to handle video-based, multi-view acquisitions. For video processing, each acquisition was decomposed into constituent frames, with each frame divided into patches processed by the vision encoder. Patch embeddings were fed sequentially to leverage the language model's positional encoding for temporal representation. The cross-view fusion mechanism synthesized information across echocardiographic views (e.g., parasternal long-axis, apical four-chamber) or CMR planes (e.g., short-axis stack, long-axis views) to enable comprehensive cardiac assessment (Supplementary Figure~3).

Training mirrored the two-stage approach used for ECG: initial training of the visual components with frozen language model weights, followed by joint fine-tuning of all parameters. Language model outputs were further optimized using Group Relative Policy Optimization (GRPO). Models were trained on interleaved $\langle$question$\rangle$$\langle$video$\rangle$$\langle$answer$\rangle$ sequences spanning multiple-choice questions, quantitative measurements, and free-text interpretive reasoning (Supplementary Figure~3).

\subsection{Model architecture: Multimodality}

We adopted a modular agentic approach for multimodal integration, wherein modality-specific models communicate through natural language rather than shared embedding spaces. This architectural decision was motivated by two considerations. First, the rapid pace of large language model development (with new state-of-the-art models released approximately every three months) necessitates future-proof architectures that can incorporate advances without extensive retraining. Second, while embedding-space approaches (e.g., NExT-GPT, LLaVA-NeXT) train projection layers to map each modality's representations into a unified token space, each new language model introduces a different embedding space and tokenizer, requiring projection layers to be retrained for compatibility. By using natural language as the interface between modality-specific models, the agentic framework enables seamless integration of updated or alternative language models without retraining visual encoders or projection modules. Third, the text-based orchestrator reduces mirages we have observed in other vision language models, as described in our companion paper (Mirage)\citep{asadi2026mirage}.

The orchestrator additionally implements a confidence-assessment protocol designed to detect outputs not grounded in the provided imaging data. To guard against mirage reasoning, the orchestrator subjects each modality expert to a three-step verification procedure at inference time. First, for each clinical sub-query, the orchestrator generates three semantically equivalent but syntactically distinct rephrasings and routes all variants to the relevant expert model. Second, the orchestrator issues an image-absent version (mirage-mode) of the same query as a counterfactual probe that establishes the expert model's language prior without access to any visual input. Third, a per-modality confidence score is computed from two signals: the consistency of the expert's responses across the three rephrased queries, and the divergence between the image-present responses and the image-absent baseline. High inter-rephrase consistency coupled with high divergence from the counterfactual baseline indicates visually grounded reasoning; conversely, responses that closely resemble the image-absent output, regardless of their apparent confidence, are flagged as potentially mirage-affected. The orchestrator weights each modality's contribution to the final assessment proportionally to its confidence score, and communicates residual uncertainty to the user when one or more modalities are flagged.

\subsection{Evaluation}

MARCUS was evaluated across two complementary paradigms: visual multiple-choice accuracy and open-ended clinical reasoning quality. For multiple-choice evaluation, each modality-specific model and the multimodal agent were tested on a held-out test question set requiring selection of a single correct answer from 4--5 options. Accuracy was calculated as the proportion of correctly answered questions. For open-ended evaluation, models generated free-text responses to clinical queries, which were compared against ground-truth physician reports using Likert scores (1--5 scale). Likert adjudication was performed by a blinded expert cardiologist for a subset of questions and by a separate large language model for the complete dataset.

Performance was benchmarked against two large frontier vision language models: GPT-5 (Thinking) and Gemini 2.5 Pro (Deep Think). All models received identical inputs and questions. Statistical comparisons between MARCUS and comparator models used two-sided paired tests with significance defined as $P<0.05$.

Internal validation was performed on held-out Stanford data not used during training. External validation was conducted on an independent cohort from UCSF, comprising distinct patients, imaging equipment, and acquisition protocols. No UCSF data were used during model development, and no additional fine-tuning was performed prior to external evaluation. Generalizability was assessed by comparing accuracy and Likert scores across sites, with statistical testing for between-site differences.

To ensure that our evaluation benchmarks measured genuine visual reasoning rather than exploitation of textual patterns, we applied three complementary strategies. First, each question was generated in multiple syntactically distinct rephrasings. Second, for a subset of multiple-choice questions, one answer option was replaced with ``none of the other options.'' Third, all models were evaluated in an image-absent condition. Questions that any evaluated model answered correctly without image access were excluded from the primary performance comparisons, following the B-Clean framework\citep{asadi2026mirage}. This filtered evaluation retained 60\% of the original question set.

\subsection{Benchmarking MARCUS against single-modality cardiac AI models}

To contextualise MARCUS's performance against purpose-built AI systems for cardiac imaging interpretation, we benchmarked MARCUS against four single-modality models spanning the three cardiac modalities: EchoNext\citep{poterucha2025} (12-lead ECG),
EchoPrime\citep{vukadinovic2026} and PanEcho\citep{holste2025}
(echocardiography), and ViTa\citep{zhang2025} (cardiac MRI). Each
single-modality model was applied to its intended modality on (i) the
held-out Stanford internal test set and (ii) external public test sets in
the same modality (Supplementary Methods, Section~\ref{supp:public_datasets}).

External validation used five public test sets spanning four countries:
ACDC\citep{bernard2018} (cardiac MRI; Dijon University Hospital, France;
520 MCQ, 50 VQA held-out test items), the UK Biobank imaging
substudy\citep{sudlow2015} (cardiac MRI; United Kingdom; 337 MCQ,
29 VQA held-out test items),
MIMIC-IV-ECHO\citep{johnson2023mimiciv} (transthoracic echocardiography;
Beth Israel Deaconess Medical Centre, Boston, USA; 4{,}795 MCQ, 750 VQA),
MIMIC-IV-ECG\citep{gow2023ecg} (12-lead ECG; same Boston cohort; 3{,}000
MCQ, 479 VQA), and PTB-XL\citep{wagner2020} (12-lead ECG;
Physikalisch-Technische Bundesanstalt repository, Germany; 1{,}676 MCQ,
321 VQA).

For free-text VQA evaluation, structured single-modality model outputs
(EchoNext binary probabilities, PanEcho structured predictions, ViTa
phenotype scalars) were formatted into clinical pseudo-reports (EchoPrime natively generates text reports) and graded
against the narrative ground-truth answer by an independent LLM-as-judge
on the same 1--5 Likert rubric used for MARCUS (1 = clinically wrong;
5 = fully correct; binarised at $\geq 4$ for the binary-accuracy column). 
Accuracy and free-text mean Likert were reported with 95\% confidence
intervals from 2{,}000-iteration bootstrap resampling. Paired differences
in binary correctness were tested using McNemar's exact test on the
$2 \times 2$ contingency of concordant and discordant item-level
responses; paired differences in mean Likert were tested using the
Wilcoxon signed-rank test (two-sided). Full inference protocols,
pseudo-report templates, the clinical-deployment-fairness rule, and
per-comparison sample sizes are provided in Supplementary Methods, Section~\ref{supp:single_modality}).

\section*{Acknowledgements}
We would like to thank the Knight Initiative for Brain Resilience for unrestricted funding to support GPU and computational resources. Mohammad Asadi was funded through Amazon AI Fellowship.

\section*{Author contributions}
Conceptualization: J.W.O., M.A., E.A.A. Methodology: M.A., T.N., J.W.O, E.Ad., E.A.A. Software: M.A., L.E. Formal analysis: M.A., L.E., J.W.O, T.N. Investigation: J.W.O., M.A. Data curation: J.W.O., M.A., R.A., E.A.A. Writing---original draft: J.W.O., M.A., E.A.A. Writing---review \& editing: all authors. Visualization: J.W.O., M.A., T.N. Supervision: R.A., E.A.A. Funding acquisition: E.A.A.

\section*{Competing interests}
E.A. is a Founder of Personalis, Deepcell, Svexa, Saturnus Bio, and Swift Bio, Founding Advisor of Candela, and Parameter Health, Advisor for Pacific Biosciences, and Non-executive director of AstraZeneca, and Dexcom. J.O.S. is supported by the AHA Postdoctoral fellowship and ACC postdoctoral fellowship and has had consultancy relationships with Google AI, Curie Bio, and Foresite Labs (outside the current work). A.S.C., unrelated to this work, is a co-founder and receives salary support from Cognita Imaging, has equity interest in Radiology Partners, Subtle Medical, Brain Key and LVIS Corp., and has provided consulting services to Patient Square Capital, Elucid Bioimaging, and Chondrometrics. All their work was performed as a part of Stanford University. All other authors declare no relevant conflicts of interests.

\section*{Code availability}
All source code for MARCUS is publicly available at \url{https://github.com/AshleyLab/MARCUS} under an MIT license.

\section*{Data availability}
The test set questions are available in the GitHub repository. The full MARCUS-Benchmark dataset and model weights will be released upon acceptance. Raw clinical imaging data from Stanford and UCSF are protected under institutional data use agreements and are not publicly available.

\bibliography{references}

\newpage
\begin{center}
{\LARGE \textbf{Supplementary Information}} \\[12pt]
{\large MARCUS: An Agentic, Multimodal Vision-Language Model for Cardiac Diagnosis and Management}
\end{center}
\vspace{24pt}
\setcounter{figure}{0}
\setcounter{table}{0}
\setcounter{section}{0}
\renewcommand{\figurename}{Supplementary Figure}
\renewcommand{\tablename}{Supplementary Table}
\maketitle

\section{Stanford Data Question Test Set}
\label{supp:stanford_test_set}

\subsection{Dataset Overview}

Model performance was primarily evaluated on a held-out Stanford test set comprising cardiac imaging studies across four modalities: transthoracic echocardiography (Echo), cardiac magnetic resonance imaging (CMR), electrocardiography (ECG), and a multimodal task combining two or more imaging types. Three models were evaluated: MARCUS, Gemini 2.5 Pro (Deep Think), and GPT-5 (Thinking).

\subsection{Dataset Composition}

The Stanford test set included both open-ended visual question answering (VQA) and multiple-choice question (MCQ) formats (Table~S1).

VQA Questions: 100 items per single modality (drawn from 78--82 unique clinical studies).

MCQ Questions: Single-Modality: 50 items per modality (Echo, CMR, ECG) from 41--46 unique studies. Multimodal: 40 items pairing Echo and CMR studies.

The multimodal task required models to integrate information from paired imaging of two modalities simultaneously to answer a single question.

\begin{table}[H]
\centering
\caption{Stanford test set composition by modality and question type.}
\label{tab:stanford}
\begin{tabular}{lcccc}
\toprule
Modality & VQA Questions & VQA Unique Studies & MCQ Questions & MCQ Unique Studies \\
\midrule
Echo & 100 & 82 & 50 & 46 \\
CMR & 100 & 78 & 50 & 46 \\
ECG & 100 & 82 & 50 & 41 \\
Multimodal & 100 & --- & 40 & --- \\
\midrule
Total & 400 & & 190 & \\
\bottomrule
\end{tabular}
\end{table}

\subsection{Question Categories}

VQA questions were assigned to clinically relevant categories per modality. Echo categories included anatomy, valves, measurements, and function. CMR categories included function, valves, anatomy, measurements, and tissue characterisation. ECG categories included wave morphology, rhythm, ischaemia/infarction, and voltage/hypertrophy/strain. Multimodal questions were grouped into seven higher-level categories: ventricular function, valve morphology, valvular regurgitation, chamber anatomy, flow obstruction, great vessels, and pericardium.

\subsection{Model Evaluation}

All three models were evaluated on identical question sets. VQA responses were scored on a 1--5 Likert scale using an automated evaluation framework. MCQ accuracy was computed by extracting the predicted answer letter and comparing it to the ground-truth answer using regular expression matching.

\section{UCSF Question External Validation Dataset}
\label{supp:ucsf_test_set}

\subsection{Dataset Overview}

To evaluate external generalisability, MARCUS and GPT-5 were assessed on an independent dataset derived from the University of California, San Francisco (UCSF). The UCSF validation dataset comprised cardiac imaging studies across three modalities: transthoracic echocardiography (Echo), cardiac magnetic resonance imaging (CMR), and electrocardiography (ECG). All study content including questions, ground-truth answers, and model responses was anonymised prior to analysis.

\subsection{Dataset Composition}

The full UCSF dataset encompassed approximately 400--460 unique clinical studies per modality (Table~S2). From each study, multiple questions were generated spanning both open-ended visual question answering (VQA) and multiple-choice question (MCQ) formats, yielding a total of 19,841 VQA questions and 19,262 MCQ questions across all three modalities.

\begin{table}[H]
\centering
\caption{UCSF dataset composition by modality and question type.}
\label{tab:ucsf}
\begin{tabular}{lccccc}
\toprule
Modality & Unique Studies & VQA Questions & MCQ Questions & Avg VQA Q/study & Avg MCQ Q/study \\
\midrule
Echo & 411--414 & 8,119 & 10,345 & 19.8 & 25.0 \\
CMR & 455--457 & 6,874 & 6,226 & 15.0 & 13.7 \\
ECG & 406--409 & 4,572 & 2,691 & 11.3 & 6.6 \\
\midrule
Total & ${\sim}$1,274 & 19,565 & 19,262 & & \\
\bottomrule
\end{tabular}
\end{table}

\subsection{Model Evaluation}

MARCUS was evaluated on a randomly sampled subset of 1,000 questions per modality for both VQA and MCQ tasks, drawn from the full UCSF dataset. Performance on the complete dataset was additionally computed for MCQ (Echo: $n$=10,345; CMR: $n$=6,226; ECG: $n$=2,691). GPT-5 (gpt-5-2025-08-07) was evaluated on a 200-question subset per modality for both task types. Gemini 2.5 Pro was not evaluated on the UCSF dataset. VQA responses were scored on a 1--5 Likert scale using the same automated evaluation framework applied to the Stanford test set. MCQ accuracy was computed by extracting the predicted answer letter and comparing it to the ground-truth answer using regular expression matching.

\section{Public datasets for external validation}
\label{supp:public_datasets}

To complement the held-out Stanford internal test set (Supplementary Section~\ref{supp:stanford_test_set}) and
the UCSF external validation cohort ((Supplementary Section~\ref{supp:ucsf_test_set})), we curated MCQ and
free-text VQA test items from five publicly available cardiac imaging
datasets spanning three modalities and four countries (Supplementary
Table~\ref{tab:public_datasets}). Question generation followed the same
two-stage process used for Stanford and UCSF: an expert cardiologist
developed a template library of clinically relevant question types,
and each dataset's accompanying clinical text (physician interpretation
report, structured diagnostic label, or expert-validated annotation) was
processed by a large language model to generate dataset-specific
questions, with ground-truth answers extracted directly from the
accompanying text. For multiple-choice items, the expert cardiologist
reviewed each question and created plausible distractor options. 

\begin{table}[H]
\centering
\caption{\textbf{Public-dataset external validation cohort.} Source
modality, country of origin, and item counts per dataset.}
\label{tab:public_datasets}
\footnotesize
\begin{tabular}{@{}lllrr@{}}
\toprule
\textbf{Dataset} & \textbf{Modality} & \textbf{Country} &
\textbf{$n$ MCQ} & \textbf{$n$ VQA} \\
\midrule
ACDC                & Cardiac MRI & France  & 520     & 50  \\
UK Biobank          & Cardiac MRI & UK      & 337     & 29  \\
MIMIC-IV-ECHO       & Echo        & USA     & 4{,}795 & 750 \\
MIMIC-IV-ECG        & 12-lead ECG & USA     & 3{,}000 & 479 \\
PTB-XL              & 12-lead ECG & Germany & 1{,}676 & 321 \\
\bottomrule
\end{tabular}
\end{table}

\paragraph{ACDC.} The Automated Cardiac Diagnosis Challenge\citep{bernard2018}
contributed 100 cine cardiac-MRI studies acquired at Dijon University
Hospital, France, with end-diastolic and end-systolic short-axis (SAX)
and long-axis (LAX) cines and expert-validated diagnostic labels
(normal, dilated cardiomyopathy, hypertrophic cardiomyopathy, prior
myocardial infarction, abnormal RV) used for question generation.

\paragraph{UK Biobank.} A randomly sampled subset of cine cardiac-MRI
studies from the UK Biobank imaging substudy\citep{sudlow2015}, with
the standard short-axis stack and long-axis (2-, 3-, 4-chamber) cine
acquisitions and accompanying derived phenotype values (LV/RV ejection
fraction, LV mass, indexed volumes) used for question generation.

\paragraph{MIMIC-IV-ECHO.} Transthoracic echocardiographic studies from
Beth Israel Deaconess Medical Centre, Boston\citep{johnson2023mimiciv}, with paired original clinical interpretation report.
Reports were used as the ground-truth source for question generation,
following the same pipeline used for Stanford echo data.

\paragraph{MIMIC-IV-ECG.} 12-lead ECG waveforms (500~Hz, 10~s) from the
same MIMIC-IV cohort\citep{gow2023ecg}, paired with their clinical
interpretation reports. Waveforms were rendered to the same $4 \times 3$
image grid format used for Stanford ECGs (Methods, Section~``ECG,
echocardiogram and CMR data processing'') for input to MARCUS, and
passed in raw waveform form to EchoNext.

\paragraph{PTB-XL.} 12-lead ECGs from the Physikalisch-Technische
Bundesanstalt repository, Germany\citep{wagner2020}, paired with
cardiologist-validated structured diagnostic labels (SCP-ECG codes
spanning rhythm, conduction, hypertrophy, ischaemia, and clinical
context) and free-text statements. Both the structured labels and the
free-text statements were used as ground-truth sources for question
generation. Waveforms were rendered to the same $4 \times 3$ image grid
format used elsewhere.

\section{MARCUS Architecture and Training Details}

\subsection{Hyperparameter Table}

The hyperparameters used in each of the 3 stages are detailed in Supplementary Table~S3.

\begin{table}[H]
\centering
\caption{Training hyperparameters for all three stages of the MARCUS training pipeline.}
\label{tab:hyperparams}
\footnotesize
\begin{tabular}{lccc}
\toprule
Parameter & Stage 1 (Pretrain) & Stage 2 (SFT) & Stage 3 (GRPO) \\
\midrule
Base model & Qwen2.5-VL-3B-Instruct & Qwen2.5-VL-3B-Instruct & Qwen2.5-VL-3B-Instruct \\
Trainable & Vision encoder + MLP & Full model & Full model (policy) \\
Frozen & Qwen2 LLM & None & Reference model \\
Learning rate & 2.0e-4 & 1.0e-5 & 1.0e-6 \\
LR scheduler & Cosine & Cosine & --- \\
Warmup ratio & 0.05 & 0.05 & --- \\
Epochs & 3 & 5 & 15 \\
Batch size /device & 8 (ECG) & 4 (ECG), 2 (Echo/CMR) & 32 (train), 16 (val) \\
Grad.\ accumulation & 4 & 8 & --- \\
Optimizer & AdamW & AdamW & AdamW \\
Weight decay & 0.0 & 0.01 & --- \\
Max grad norm & 1.0 & 1.0 & --- \\
Precision & BF16 + TF32 & BF16 + TF32 & BF16 \\
Max tokens & 1,024 & 2,048 & 5,000 + 512 \\
Validation split & --- & 1\% & Separate parquet \\
GRPO group size & --- & --- & 4 \\
KL loss coeff.\ & --- & --- & 0.01 (low\_var\_kl) \\
Entropy coeff.\ & --- & --- & 0 \\
Rollout backend & --- & --- & sglang \\
\bottomrule
\end{tabular}
\end{table}

\subsection{Hardware and Training Times}

All training was performed on NVIDIA DGX H100 systems. Stage 1 (vision encoder pretraining) used 4 H100 80\,GB GPUs for ECG (${\sim}$12 hours) and 8 H100 80\,GB GPUs for Echo (${\sim}$48 hours) and CMR (${\sim}$72 hours). Stage 2 (SFT) used 8 H100 80\,GB GPUs per modality (${\sim}$24 hours each). Stage 3 (GRPO) used 8 H100 80\,GB GPUs per modality (${\sim}$36 hours each, 15 epochs). All models used Fully Sharded Data Parallel (FSDP) for distributed training. Training was launched via \texttt{torchrun --nproc\_per\_node=8} for Stages 1--2 (LLaMA-Factory) and via the verl trainer with sglang rollout backend for Stage 3.

\subsection{GRPO Reward Function}

Group Relative Policy Optimization (GRPO) was implemented using the verl framework with a binary correctness reward. For each training prompt, the policy generates a group of $n$=4 candidate responses. Each response is scored by a deterministic reward function: 1.0 if the extracted MCQ answer letter matches the ground-truth answer, 0.0 otherwise. Answer extraction uses a cascade of regular expression patterns that match common response formats (e.g., ``The answer is B'', ``B.\ Mild LV dilation'', standalone letter). The group baseline is estimated as the mean reward across the 4 responses, and the policy gradient is computed relative to this baseline using a clipped ratio objective analogous to PPO but without a learned value function. A low-variance KL divergence penalty (coefficient 0.01) regularises the updated policy against the frozen reference model to prevent reward hacking and mode collapse. Unlike standard PPO, GRPO eliminates the need for a separate critic network, reducing computational cost while maintaining stable training dynamics.

\subsection{Mirage Analysis}

The counterfactual mirage probing protocol operates as follows. For each clinical sub-query, the orchestrator generates three semantically equivalent but syntactically distinct rephrasings using predefined templates (e.g., direct question form: ``\{question\}''; descriptive form: ``Please describe \{subject\} as seen in the provided \{modality\} data.''; imperative form: ``Based on the \{modality\} provided, answer the following: \{question\}''). All three rephrasings are sent to the relevant expert model together with the image or video. The orchestrator then sends the original query without any visual input as a counterfactual probe.

Two scores are computed: (1) inter-rephrase consistency, defined as the mean pairwise Jaccard token similarity across the three image-present responses; and (2) image-absent divergence, defined as 1 minus the mean Jaccard similarity between the image-present responses and the image-absent response. A response is flagged as a potential mirage when the image-absent similarity exceeds a threshold of 0.85 (i.e., divergence $<$ 0.15), indicating that the model's output did not meaningfully change when visual input was removed. Composite confidence is computed as the mean of consistency and divergence scores, equally weighted.

\section{Single-modality cardiac AI model inference protocols}
\label{supp:single_modality}

\subsection{Clinical-deployment-fairness rule and scope of evaluation}

MARCUS was designed as a generalist multimodal model intended for
deployment across the full breadth of cardiac queries that
arise in clinical practice — from binary diagnostic recognition to
complex, multi-finding interpretive reasoning across all three
modalities. Some of the four single-modality models against which we benchmarked
MARCUS were designed for narrower, discrete clinical use cases:
EchoNext\citep{poterucha2025} for detecting twelve specific
structural-heart-disease labels from the ECG, and ViTa\citep{zhang2025} for
predicting eighteen quantitative cardiac phenotypes from cine MRI. Both
EchoPrime\citep{vukadinovic2026} and PanEcho\citep{holste2025} were designed for generating structured echocardiographic reports and pre-defined
parameters. Because clinicians cannot pre-filter their questions to fall
within any single model's training target, real-world deployment of a
narrow single-modality model requires it to attempt every clinical
question that arises; a model that silently abstains on out-of-scope
queries is not a clinically deployable substitute for a generalist.
We therefore evaluated single-modality models under a
\textit{clinical-deployment-fairness rule}\citep{kelly2019key,wiens2019nodono}: on the external public test
sets, questions outside the single-modality model's
training target or output scope were not excluded and the model was asked to generate an answer, to ensure fair comparison with MARCUS. This reflects what clinician would observe if the model were deployed
as a standalone interpretation tool against the full diagnostic
question stream of routine clinical practice.

For the Stanford internal evaluation only, we additionally performed
an in-scope paired comparison restricted to the subset of items
falling within each single-modality model's training target (e.g., for
EchoNext, the 153 MCQ and 62 VQA items mapping to one of its twelve
labels; see Section~\ref{supp:echonext}). This in-scope analysis is
a strictly more favourable comparison for the single-modality model
than the deployment-fairness rule and gives the most charitable
estimate of its peak achievable performance against MARCUS.

\subsection{EchoNext (ECG)}
\label{supp:echonext}

EchoNext\citep{poterucha2025} is a 1D residual convolutional neural
network trained on over one million paired ECG--echocardiogram studies
to detect structural heart disease from 12-lead ECG waveforms. The
released checkpoint produces twelve independent sigmoid probabilities
(LVEF $\leq 45$\%, LV wall thickness $\geq 1.3$~cm, moderate-or-greater
aortic stenosis, aortic regurgitation, mitral regurgitation, tricuspid
regurgitation and pulmonary regurgitation, moderate-or-greater
right-ventricular systolic dysfunction, moderate-or-large pericardial
effusion, pulmonary-artery systolic pressure $\geq 45$~mmHg, peak
tricuspid regurgitation velocity $\geq 3.2$~m/s, and a composite
structural-heart-disease label).

ECGs (500~Hz, 12-lead, 10~s) were preprocessed following the authors'
published pipeline: naive downsampling to 250~Hz, re-derivation of
leads III, aVR, aVL, and aVF from the independent leads I and II,
two-stage median-filter baseline-wander removal (kernel lengths 0.2
and 0.6~s), per-lead clipping to the authors' published bounds, and
per-lead $z$-score normalisation. 

\textit{Stanford internal evaluation (in-scope paired comparison).}
Of 3{,}516 Stanford ECG test items (1{,}647 MCQ; 1{,}869 VQA), 247
items (171 MCQ; 76 VQA; 7.0\% of the full set) fell within EchoNext's
12-label scope, ascertained by keyword-matching the question text
against each label's name. The remaining items queried rhythm,
conduction, axis, wave morphology, ischaemia, device status, or
quantitative measurements outside EchoNext's training target. For each
in-scope MCQ, EchoNext's probability for the relevant label was
thresholded at 0.5 and mapped deterministically to one of the four
answer options: a positive prediction to the option describing the
EchoNext finding at its training-threshold severity, and a negative
prediction to the option describing absence. Items in which neither
pole matched any answer option were excluded, ensuring EchoNext was
never awarded a correct answer by ambiguous mapping. For in-scope
free-text items, the ground-truth answer was parsed into a binary
finding-present/absent label by keyword classification. The Stanford
paired comparison was restricted to the resulting in-scope subset of
$n=153$ MCQ items and $n=62$ VQA items; MARCUS was scored on the
identical items.

\textit{External evaluation (clinical-deployment-fairness rule).}
EchoNext was applied to the full MIMIC-IV-ECG ($n=3{,}000$ MCQ;
$n=479$ VQA) and PTB-XL ($n=1{,}676$ MCQ; $n=321$
VQA) test sets with the same preprocessing and
zero-tabular policy. For items inside EchoNext's 12-label scope,
threshold-and-map scoring was applied as for Stanford; items outside
that scope were marked incorrect under the clinical-deployment-fairness
rule. MARCUS was scored on the same full test sets.

\textit{Mean-Likert pseudo-report grading.} For free-text mean-Likert
comparisons, EchoNext's twelve sigmoid probabilities were formatted as
a structured clinical pseudo-report listing each predicted finding
(positive or negative at the 0.5 threshold) with its associated
probability and an explicit out-of-scope notice for findings beyond the
12-label set. The pseudo-report was then graded against the narrative
ground-truth answer by the same LLM-as-judge used for MARCUS on the
1--5 Likert rubric (1 = clinically wrong; 5 = fully correct). Paired
Wilcoxon comparisons used $n=62$ items on Stanford, $n=263$ items on
the MIMIC-IV-ECG held-out subset, and $n=237$ items on the PTB-XL
held-out subset.

\subsection{EchoPrime (echocardiography)}

EchoPrime\citep{vukadinovic2026} is a multi-video vision-language
foundation model that jointly encodes all cine clips of an
echocardiographic study and synthesises a structured free-text report
via contrastive retrieval from a labelled candidate corpus.

\textit{Stanford internal evaluation.} For each of the 117 unique
Stanford echo studies, all available cine MP4 clips (median 43 per
study, range 20--60) were passed through EchoPrime's published
inference pipeline: view classification, per-clip spatio-temporal
encoding, aggregation into a study embedding, and retrieval-based
report generation. This matched the signal available to MARCUS, which
receives all views of the same study tiled into a multi-view video
grid.

Because EchoPrime outputs a free-text report rather than discrete
probabilities, question scoring used a two-stage LLM-as-judge pipeline
with GPT-4o at temperature 0. For each MCQ, GPT-4o was given the
EchoPrime report and the four answer options and returned the single
letter best supported by the report. For each VQA, GPT-4o produced a
one-sentence answer derived from the report, and a second judge pass
scored the answer against ground truth on the 1--5 Likert rubric
(binarised at $\geq 4$). Stanford paired comparisons used $n=9{,}008$
MCQ items and $n=4{,}980$ VQA items.

\textit{External evaluation.} EchoPrime was applied to MIMIC-IV-ECHO
studies and scored on $n=4{,}795$ MCQ items and $n=750$ VQA items
using the same pipeline.

\subsection{PanEcho (echocardiography)}

PanEcho\citep{holste2025} produces 40 structured predictions per
echocardiographic study spanning quantitative cardiac indices and
qualitative findings, with dedicated regression heads for several
Doppler-derived measurements (E/e$'$, RVSP, AV peak velocity, TV peak
gradient). Inference used the authors' published pipeline on the same
echo studies and MP4 clips as EchoPrime. PanEcho's structured
predictions were formatted as a clinical pseudo-report and passed
through the same two-stage GPT-4o judge pipeline used for EchoPrime.
Stanford paired comparisons used $n=9{,}017$ MCQ and $n=3{,}088$ VQA
items; external MIMIC-IV-ECHO paired comparisons used $n=4{,}795$
MCQ and $n=750$ VQA items.

\subsection{ViTa (cardiac MRI)}

ViTa\citep{zhang2025} is a 3D+T vision-transformer foundation model
pretrained on UK Biobank cardiac MRI; its published regression heads
predict 18 quantitative cardiac phenotypes (10 ventricular from SAX
cines, 8 atrial from LAX cines).

\textit{Stanford internal evaluation.} Each Stanford CMR study (200
subjects) was reformatted to ViTa's UK Biobank input specification
($128 \times 128$ spatial resolution; six apex-to-base SAX slices;
twenty-five temporal frames; three LAX views stacked as 2-/3-/4-chamber).
View classification used regular-expression matching against
\texttt{SeriesDescription}. SAX stacks with fewer than
six unique \texttt{SliceLocation}s were zero-padded by slice repetition;
missing LAX views were zero-filled. ViTa's tabular covariate branch was
fed the zero vector to enforce the same zero-tabular fair-comparison
policy applied to EchoNext. Of 200 studies, 181 were successfully
processed; 19 failed preprocessing (no cine series classifying as SAX).
ViTa's 18 phenotype predictions were formatted as a quantitative
pseudo-report with clinically calibrated normal-range annotations and
an explicit out-of-scope notice; the pseudo-report was passed through
the LLM-as-judge for MCQ scoring (single-letter answer best supported
by the report) and free-text Likert grading (1--5 rubric). Stanford
paired comparisons used $n=3{,}357$ MCQ and $n=2{,}705$ VQA items.

\textit{External evaluation.} ViTa was applied to ACDC and UK Biobank
studies under the same input-formatting protocol. External paired
comparisons used $n=520$ (ACDC MCQ), $n=50$ (ACDC VQA held-out subset),
$n=337$ (UK Biobank MCQ), and $n=29$ (UK Biobank VQA held-out subset)
items.

\subsection{MARCUS evaluation}

MARCUS was evaluated on identical items to each single-modality model,
following the same scoring pipeline for both the Stanford internal
and external evaluations described above.

\subsection{Sample sizes for paired single-modality model comparisons}

\begin{table}[H]
\centering
\caption{\textbf{Paired sample sizes for every MARCUS-versus-single-modality-model
comparison reported in the main text.} Stanford ECG sample sizes are
restricted to the in-scope paired subset (in-scope items only; see
Section~\ref{supp:echonext}); external sample sizes use the full test
set with EchoNext's out-of-scope items marked incorrect under the
clinical-deployment-fairness rule.}
\label{tab:single_modality_n}
\footnotesize
\begin{tabular}{@{}lllrrr@{}}
\toprule
\textbf{Cohort} & \textbf{Modality / dataset} & \textbf{Single-modality model} &
\textbf{$n$ MCQ} & \textbf{$n$ VQA (binary)} & \textbf{$n$ VQA (mean Likert)} \\
\midrule
\multicolumn{6}{l}{\textit{Stanford internal test set}} \\
Stanford & ECG (in-scope)    & EchoNext  & 153   & 62    & 62 \\
Stanford & Echo              & EchoPrime & 9{,}008 & 4{,}980 & 4{,}980 \\
Stanford & Echo              & PanEcho   & 9{,}017 & 3{,}088 & 3{,}088 \\
Stanford & CMR               & ViTa      & 3{,}357 & 2{,}705 & 2{,}705 \\
\midrule
\multicolumn{6}{l}{\textit{External multi-country public test sets}} \\
External & MIMIC-IV-ECG (USA)         & EchoNext  & 3{,}000 & 479 & 263 \\
External & PTB-XL (Germany)           & EchoNext  & 1{,}676 & 321 & 237 \\
External & MIMIC-IV-ECHO (USA)        & EchoPrime & 4{,}795 & 750 & 750 \\
External & MIMIC-IV-ECHO (USA)        & PanEcho   & 4{,}795 & 750 & 750 \\
External & ACDC (France)              & ViTa      & 520     & 50  & 50  \\
External & UK Biobank (UK)            & ViTa      & 337     & 29  & 29  \\
\bottomrule
\end{tabular}
\end{table}

\section{Supplementary Figures}

\begin{figure}[H]
\centering
\includegraphics[width=\textwidth]{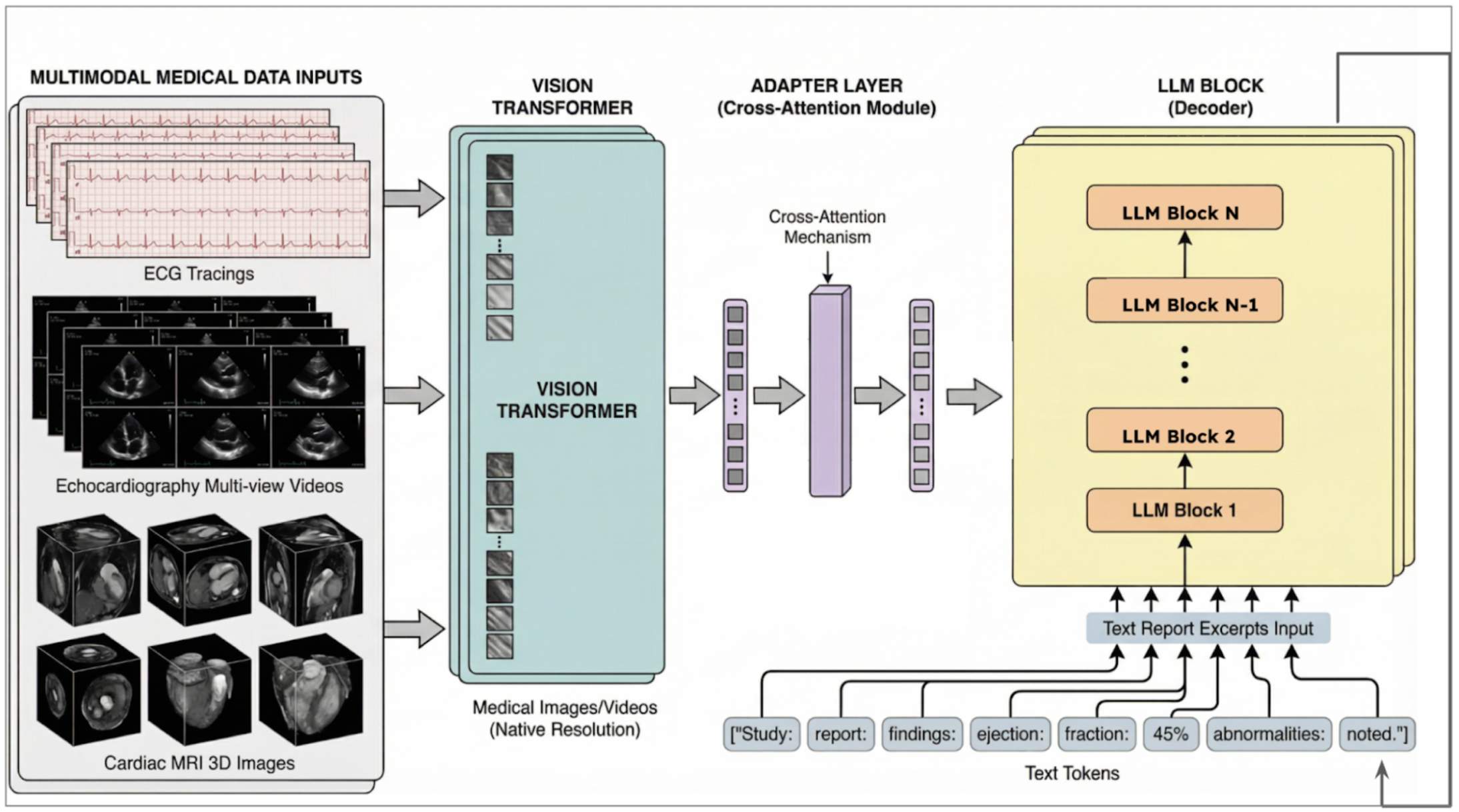}
\caption{\textbf{An overview of the vision-language model for each modality expert.} Each ECG trace, echocardiography video, or MRI video is broken into $16 \times 16$ patches and fed into the vision transformer. The vision embeddings are then run through a cross-attention layer and then integrated into the LLM blocks by (1) concatenation with the text tokens and (2) residual connections between various depths of the vision transformers (ViT) and the LLM blocks.}
\label{fig:supp1}
\end{figure}

\begin{figure}[H]
\centering
\includegraphics[width=\textwidth]{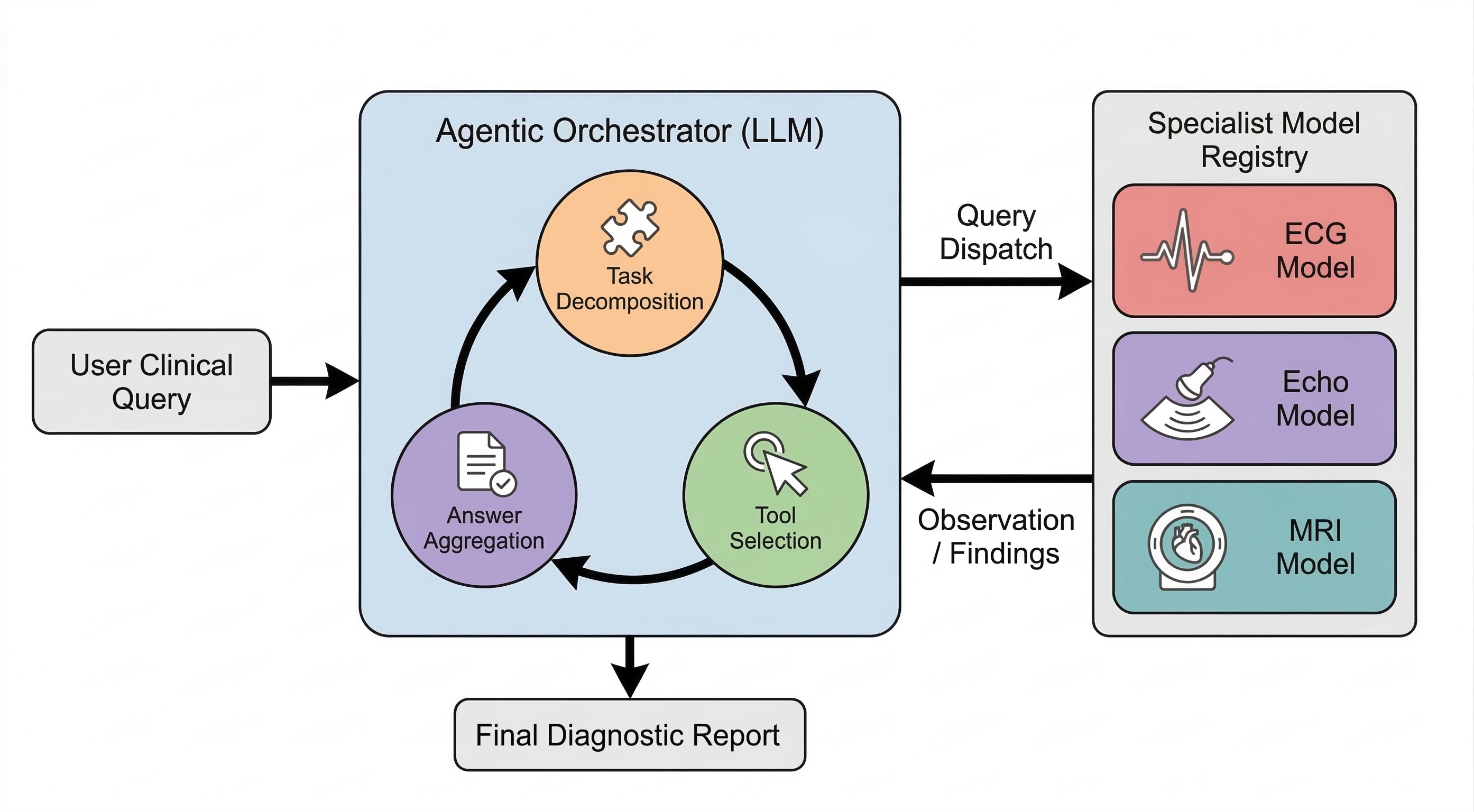}
\caption{\textbf{Agentic orchestrator architecture.} The orchestrator interacts with the user, decomposes the questions into sub-questions for each of the single modalities (if necessary), interacts with the tools (single-modality experts) to infer information from each modality, combines the observations and findings, manages conflicts, and aggregates the results into the final answer to return to the user.}
\label{fig:supp2}
\end{figure}

\begin{figure}[H]
\centering
\includegraphics[width=\textwidth]{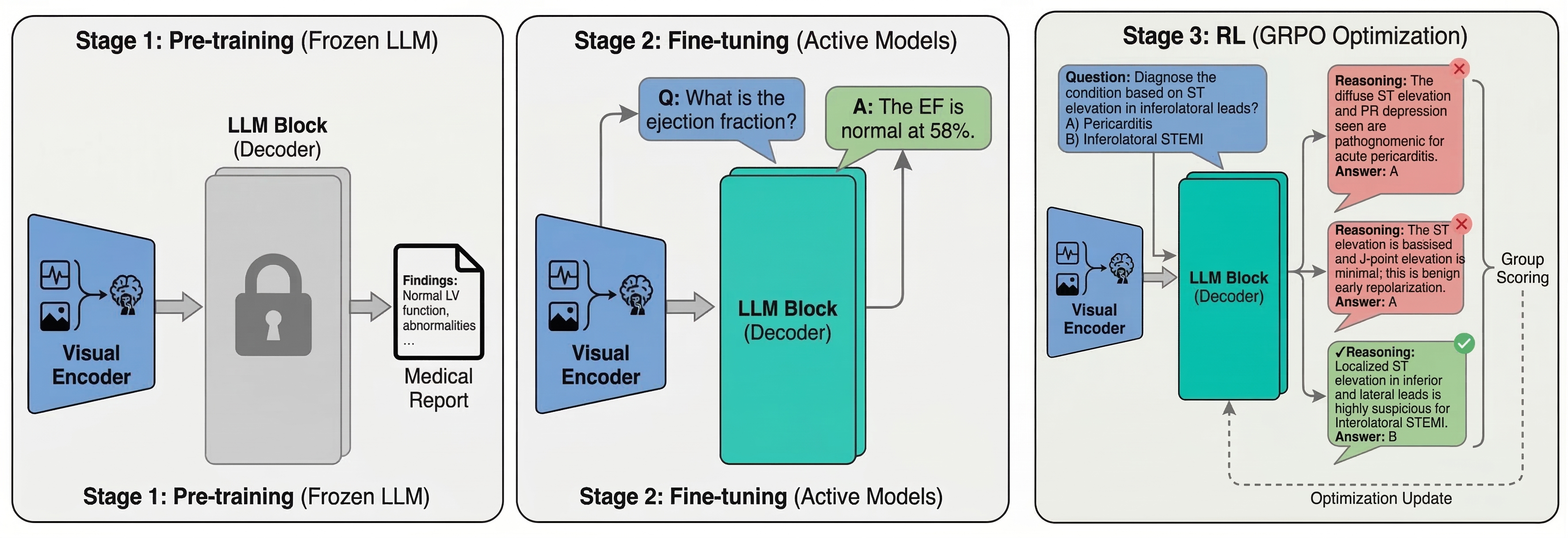}
\caption{\textbf{Multi-stage training pipeline for MARCUS.} \textbf{Stage 1:} Pre-training (Frozen LLM). A specialized visual encoder is pre-trained for distinct cardiac modalities (ECG, Echo, CMR) to extract dense latent representations from medical imaging data. During this stage, the parameters of the decoder-only large language model (LLM block) are held frozen to preserve its pre-existing linguistic knowledge while aligning the new visual features. \textbf{Stage 2:} Fine-tuning (Active Models). The pre-trained vision encoder and the LLM block are simultaneously fine-tuned on visual question answering tasks. \textbf{Stage 3:} RL (GRPO Optimization). Model precision for diagnostics-focused queries is further enhanced through reinforcement learning using Group-Relative Policy Optimization (GRPO).}
\label{fig:supp3}
\end{figure}

\begin{figure}[H]
\centering
\includegraphics[width=\textwidth]{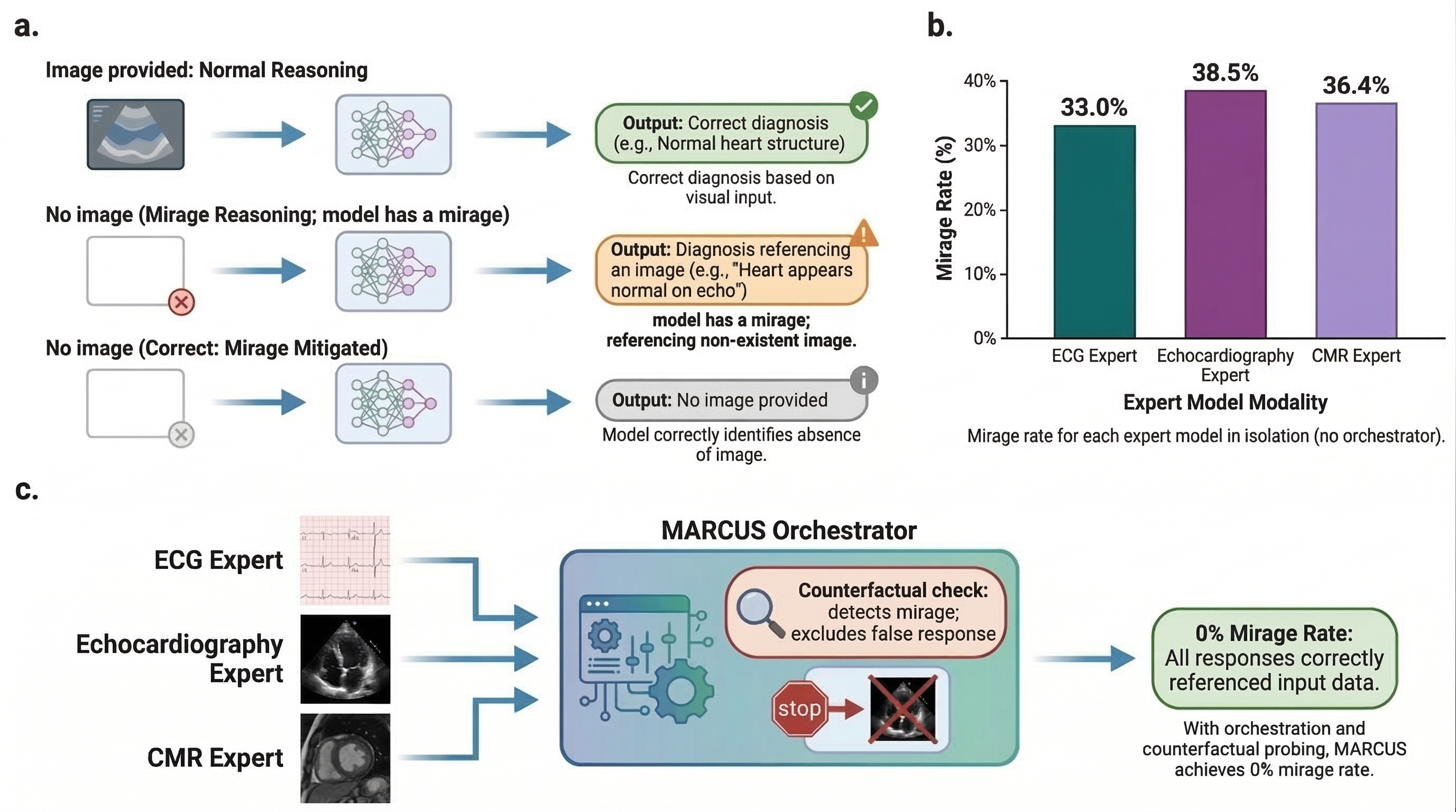}
\caption{\textbf{Susceptibility of medical expert models to mirage reasoning and its mitigation via agentic orchestration in MARCUS.} \textbf{A.}~Mechanism of mirage reasoning. \textbf{B.}~Quantification of mirage rates in modality-specific expert models within MARCUS. \textbf{C.}~System-level mitigation via the agentic Orchestrator within MARCUS achieving a zero-percent mirage rate.}
\label{fig:supp4}
\end{figure}

\section{MCQ Accuracy with 95\% CI and McNemar p-values}

Supplementary Table~S4 reports the MCQ diagnostic accuracy for each model across all four evaluation modalities, with bootstrapped 95\% confidence intervals (5,000 resamples) and exact McNemar's test p-values for pairwise comparisons between MARCUS and each frontier model. McNemar's test was selected because it accounts for the paired structure of the evaluation (all three models answered the same set of questions), providing a more statistically powerful and appropriate comparison than unpaired tests.

\begin{table}[H]
\centering
\caption{MCQ accuracy with bootstrapped 95\% confidence intervals and McNemar's exact test p-values.}
\label{tab:mcq}
\footnotesize
\begin{tabular}{llccc}
\toprule
Modality & Model & $N$ & Accuracy & 95\% CI / p-value \\
\midrule
ECG & MARCUS & 50 & 88.0\% & (78.0--96.0\%) \\
ECG & GPT-5 & 50 & 48.0\% & (34.0--62.0\%) \\
ECG & Gemini 2.5 Pro & 47 & 46.8\% & (34.0--61.7\%) \\
ECG & MARCUS vs GPT-5 (McNemar p) & 50 & --- & $<$0.001 \\
ECG & MARCUS vs Gemini (McNemar p) & 47 & --- & $<$0.001 \\
\midrule
Echo & MARCUS & 50 & 64.0\% & (50.0--78.0\%) \\
Echo & GPT-5 & 50 & 34.0\% & (22.0--48.0\%) \\
Echo & Gemini 2.5 Pro & 48 & 22.9\% & (10.4--35.4\%) \\
Echo & MARCUS vs GPT-5 (McNemar p) & 50 & --- & 0.007 \\
Echo & MARCUS vs Gemini (McNemar p) & 48 & --- & $<$0.001 \\
\midrule
CMR & MARCUS & 50 & 88.0\% & (78.0--96.0\%) \\
CMR & GPT-5 & 50 & 58.0\% & (44.0--72.0\%) \\
CMR & Gemini 2.5 Pro & 50 & 44.0\% & (30.0--58.0\%) \\
CMR & MARCUS vs GPT-5 (McNemar p) & 50 & --- & $<$0.001 \\
CMR & MARCUS vs Gemini (McNemar p) & 50 & --- & $<$0.001 \\
\midrule
Multimodal & MARCUS & 38 & 73.7\% & (60.5--86.8\%) \\
Multimodal & GPT-5 & 40 & 22.5\% & (10.0--35.0\%) \\
Multimodal & Gemini 2.5 Pro & 34 & 29.4\% & (14.7--44.1\%) \\
Multimodal & MARCUS vs GPT-5 (McNemar p) & 38 & --- & $<$0.001 \\
Multimodal & MARCUS vs Gemini (McNemar p) & 32 & --- & $<$0.001 \\
\bottomrule
\end{tabular}
\end{table}

\section{VQA Likert Scores with 95\% CI and Mann--Whitney p-values}

Supplementary Table~S5 reports mean and median Likert scores for open-ended VQA tasks, with bootstrapped 95\% confidence intervals and two-sided Mann--Whitney U test p-values for pairwise model comparisons. Mann--Whitney U was selected as the primary test because Likert scores are ordinal and their distributions were non-normal across most modality--model combinations.

\begin{table}[H]
\centering
\caption{VQA Likert score statistics.}
\label{tab:vqa}
\footnotesize
\begin{tabular}{llcccccr}
\toprule
Modality & Model & $N$ & Mean & 95\% CI (mean) & Median & 95\% CI (median) & p-value \\
\midrule
ECG & MARCUS & 100 & 3.65 & (3.37--3.91) & 4.0 & (4.0--4.0) & --- \\
ECG & GPT-5 & 100 & 2.60 & (2.34--2.87) & 3.0 & (2.0--3.0) & $<$0.001 \\
ECG & Gemini 2.5 Pro & 100 & 2.55 & (2.29--2.81) & 2.0 & (2.0--3.0) & $<$0.001 \\
\midrule
Echo & MARCUS & 100 & 2.41 & (2.14--2.69) & 2.0 & (2.0--3.0) & --- \\
Echo & GPT-5 & 100 & 1.97 & (1.75--2.20) & 2.0 & (1.0--2.0) & 0.041 \\
Echo & Gemini 2.5 Pro & 100 & 1.47 & (1.31--1.65) & 1.0 & (1.0--1.0) & $<$0.001 \\
\midrule
CMR & MARCUS & 100 & 2.91 & (2.62--3.21) & 3.0 & (2.0--3.0) & --- \\
CMR & GPT-5 & 100 & 2.19 & (1.94--2.45) & 2.0 & (1.0--2.0) & $<$0.001 \\
CMR & Gemini 2.5 Pro & 100 & 1.95 & (1.69--2.22) & 1.0 & (1.0--1.0) & $<$0.001 \\
\midrule
Multimodal & MARCUS & 100 & 3.28 & (2.92--3.64) & 4.0 & (2.0--5.0) & --- \\
Multimodal & GPT-5 & 100 & 2.69 & (2.33--3.06) & 2.0 & (1.0--3.0) & 0.036 \\
Multimodal & Gemini 2.5 Pro & 100 & 1.46 & (1.25--1.70) & 1.0 & (1.0--1.0) & $<$0.001 \\
\bottomrule
\end{tabular}
\end{table}

\section{Per-Category VQA Statistics}

Supplementary Table S6 (available as a separate CSV file: \texttt{Table\_S5\_PerCategory\_VQA\_Stats.csv}) reports mean Likert scores, 95\% bootstrap confidence intervals, and pairwise Mann--Whitney p-values for every diagnostic category within each modality. Notable entries include: ECG devices/pacing (MARCUS mean 5.00, $n$=3); ECG arrhythmia (MARCUS mean 4.50, 95\% CI 4.17--4.83, $n$=6); CMR pathology identification (MARCUS mean 3.83, 95\% CI 2.67--4.67, $n$=6); multimodal LV size (MARCUS mean 5.00, $n$=5); and multimodal wall motion pattern (MARCUS mean 1.00, $n$=7). These extremes illustrate that MARCUS performance is highly task-specific, with the most pronounced advantages in binary-like recognition tasks and the largest remaining gaps in continuous quantification tasks.

\section{MARCUS Failure Case Log}

Supplementary Table S7 (available as a separate CSV file: \texttt{Table\_S7\_MARCUS\_Failures.csv}) provides the complete log of 209 MARCUS failure cases identified across MCQ and VQA evaluation sets (40 MCQ failures; 169 VQA responses with Likert $\leq$2). This table is intended to support qualitative review by clinical readers and to facilitate future work on error-specific model improvements.

\section{Statistical Analysis}

All statistical analyses were performed using Python 3.9 with scipy 1.7, statsmodels 0.13, and numpy 1.23. For MCQ accuracy comparisons, we used McNemar's test for paired binary outcomes, which appropriately accounts for the correlated structure of the evaluation. Exact McNemar's test was used when the number of discordant pairs was fewer than 25; continuity-corrected McNemar's test was used otherwise. For VQA Likert score comparisons, we used the two-sided Mann--Whitney U test, given the ordinal nature of Likert scales and the non-normality of the distributions.

Confidence intervals for accuracy and mean Likert scores were computed using non-parametric bootstrap resampling (5,000 resamples, random seed 42) with the percentile method (2.5th and 97.5th percentiles of the bootstrap distribution). All reported p-values are two-sided. The significance threshold was set at $\alpha = 0.05$. No correction for multiple comparisons was applied to the primary modality-level comparisons, which were pre-specified.

\section{Failure Case Classification}

Failure cases were identified using two criteria: (1) for MCQ, any case where the model's extracted answer letter (A--E) did not match the ground-truth answer letter; (2) for VQA, any case assigned a Likert score of $\leq$2. Error type classification was performed using a keyword-based heuristic applied to the concatenated model response and evaluator explanation text. The four categories were: (1) Visual misinterpretation; (2) Reasoning/synthesis error; (3) Modality confusion; (4) Hallucination/fabrication. Cases not matching any keyword category were classified as Other/Unclassified.

\section{MCQ Confusion Matrices per Modality}

MARCUS demonstrated near-diagonal matrices across ECG (accuracy 88.0\%, 95\% CI 78.0--96.0\%) and CMR (88.0\%, 95\% CI 78.0--96.0\%), consistent with high-precision, low-confusion classification. Echocardiography showed a wider off-diagonal spread (accuracy 64.0\%, 95\% CI 50.0--78.0\%), reflecting the greater visual ambiguity and temporal complexity of ultrasound-based assessment. In the multimodal setting, MARCUS maintained substantially more diagonal structure (73.7\%, 95\% CI 60.5--86.8\%) compared with GPT-5 (22.5\%) and Gemini 2.5 Pro (29.4\%), whose matrices were markedly diffuse.

Both frontier models exhibited a notable over-representation in the `None of the other options' category across modalities, particularly for ECG (GPT-5: 48.0\%, Gemini: 46.8\%) and multimodal questions. This pattern is consistent with hedging behaviour in the absence of visual grounding.

\begin{figure}[H]
\centering
\includegraphics[width=\textwidth]{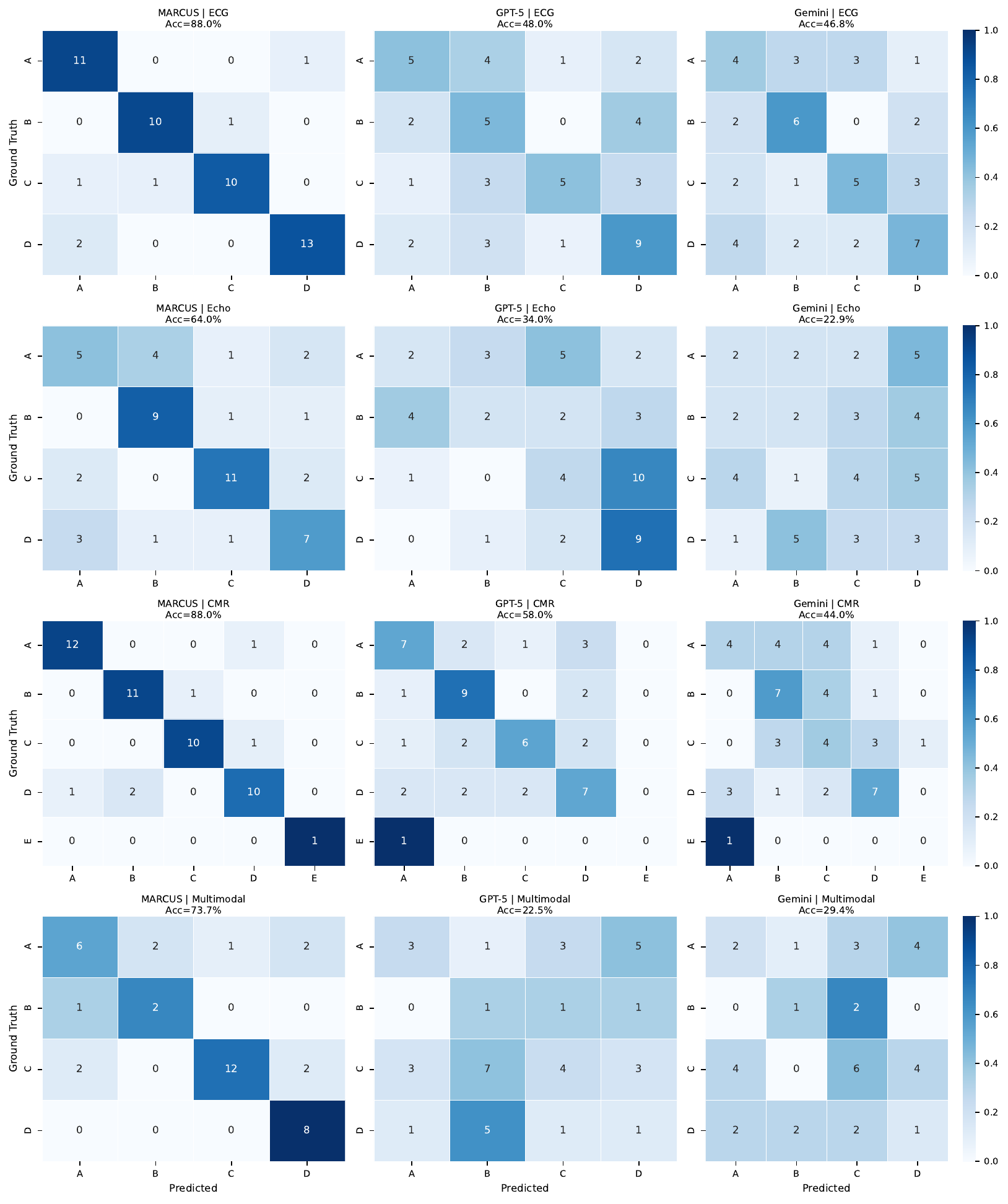}
\caption{\textbf{Confusion matrices for MCQ evaluation across modalities.} Rows represent ground-truth answer options (A--E); columns represent predicted answer options. Cell values show raw counts; colour intensity reflects the row-normalised conditional probability (0--1 scale). Diagonal cells indicate correct predictions. Three models are shown: MARCUS (left), GPT-5 (centre), Gemini 2.5 Pro (right).}
\label{fig:suppS4}
\end{figure}

\section{Per-Category VQA Likert Score Heatmaps}

Within ECG, MARCUS achieved the highest mean Likert scores in the devices/pacing category (5.00, 95\% CI 5.00--5.00; $n$=3) and arrhythmia identification (4.50, 95\% CI 4.17--4.83; $n$=6). Within echocardiography, the valves category yielded the highest MARCUS mean Likert score (3.11, 95\% CI 2.61--3.61; $n$=18). Within CMR, MARCUS performed best in pathology identification (3.83, 95\% CI 2.67--4.67; $n$=6). Within the multimodal category, MARCUS achieved the highest scores for LV size (5.00; $n$=5) and left atrial size (5.00; $n$=4). The lowest scores were observed for wall motion pattern (1.00; $n$=7) and mitral regurgitation quantification (1.29, 95\% CI 1.00--1.71; $n$=7).

\begin{figure}[H]
\centering
\includegraphics[width=\textwidth]{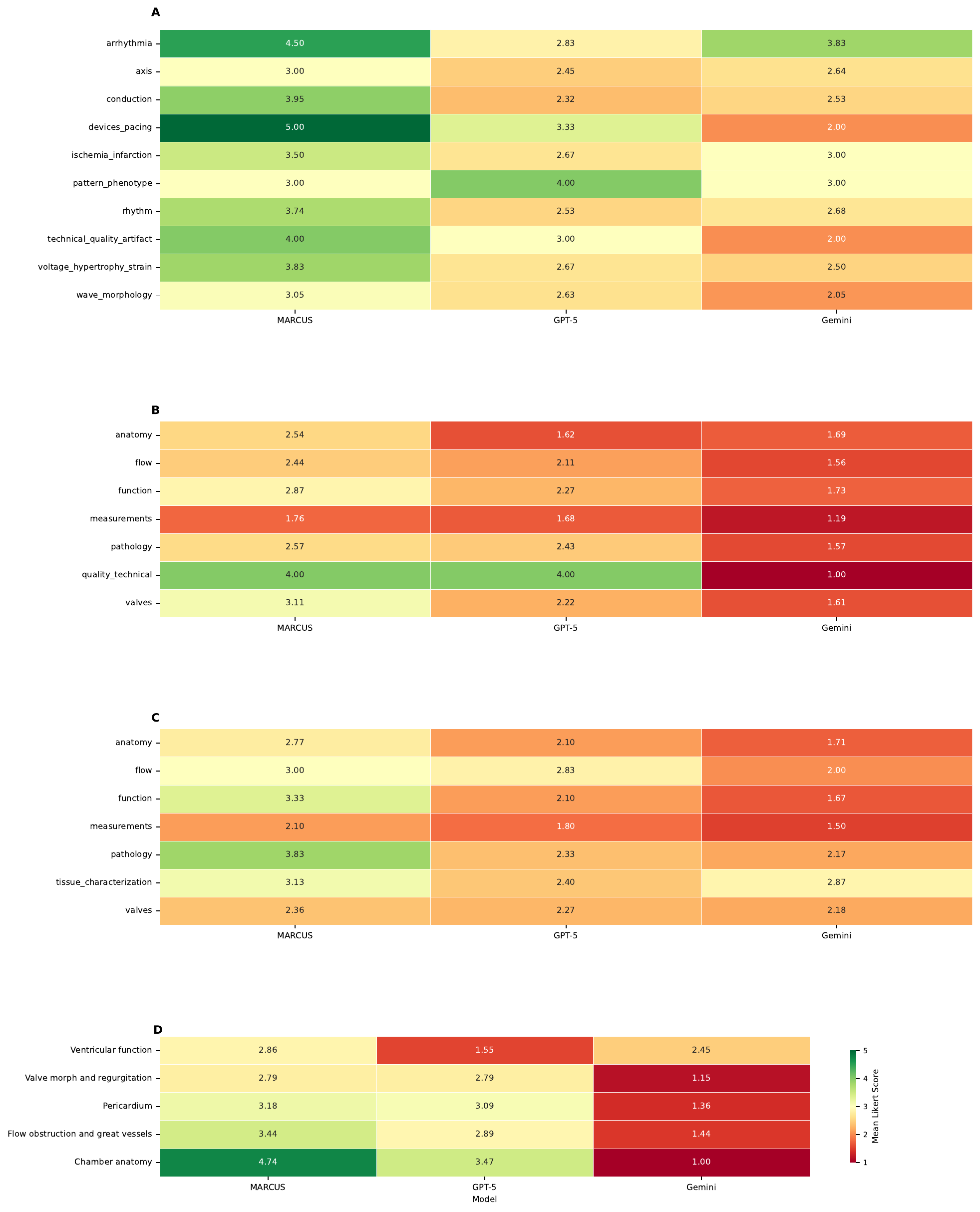}
\caption{\textbf{Per-category mean Likert score heatmaps for VQA tasks.} A: ECG categories. B: Echocardiography categories. C: CMR categories. D: Multimodal categories. Cell values are mean Likert scores (scale 1--5). Colour scale: red (1) to green (5). All three models (MARCUS, GPT-5, Gemini 2.5 Pro) are shown as column groups within each panel.}
\label{fig:suppS5}
\end{figure}

\section{Likert Score Distribution: Violin Plots}

For ECG, the MARCUS distribution was right-skewed with a median of 4.0 (mean 3.65, 95\% CI 3.37--3.91), indicating that the majority of MARCUS free-text ECG responses were rated as good to excellent. For echocardiography, all three models showed distributions skewed toward lower scores. For CMR, MARCUS showed a wider distribution (mean 2.91, 95\% CI 2.62--3.21, median 3.0). In the multimodal setting, MARCUS demonstrated a bimodal distribution (median 4.0, mean 3.28, 95\% CI 2.92--3.64).

\begin{figure}[H]
\centering
\includegraphics[width=\textwidth]{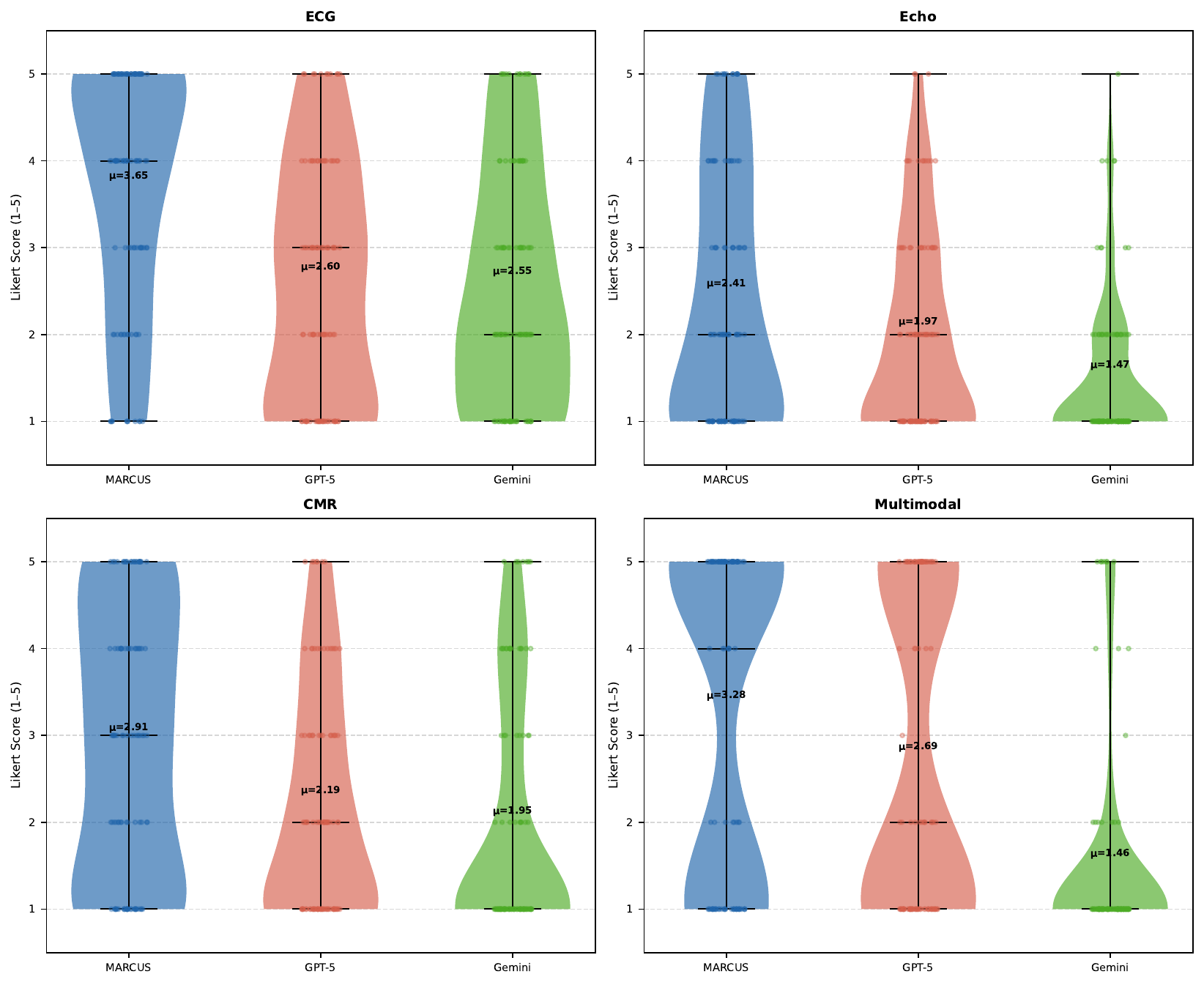}
\caption{\textbf{Violin plots of Likert score distributions per modality} for MARCUS (blue), GPT-5 (red), and Gemini 2.5 Pro (green). Individual responses are overlaid as jittered strip plots (semi-transparent points). Horizontal bars within each violin indicate the median. Annotated means ($\mu$) are shown above each violin. Likert scale: 1 = poor, 5 = excellent.}
\label{fig:suppS6}
\end{figure}

\begin{figure}[H]
\centering
\includegraphics[width=\textwidth]{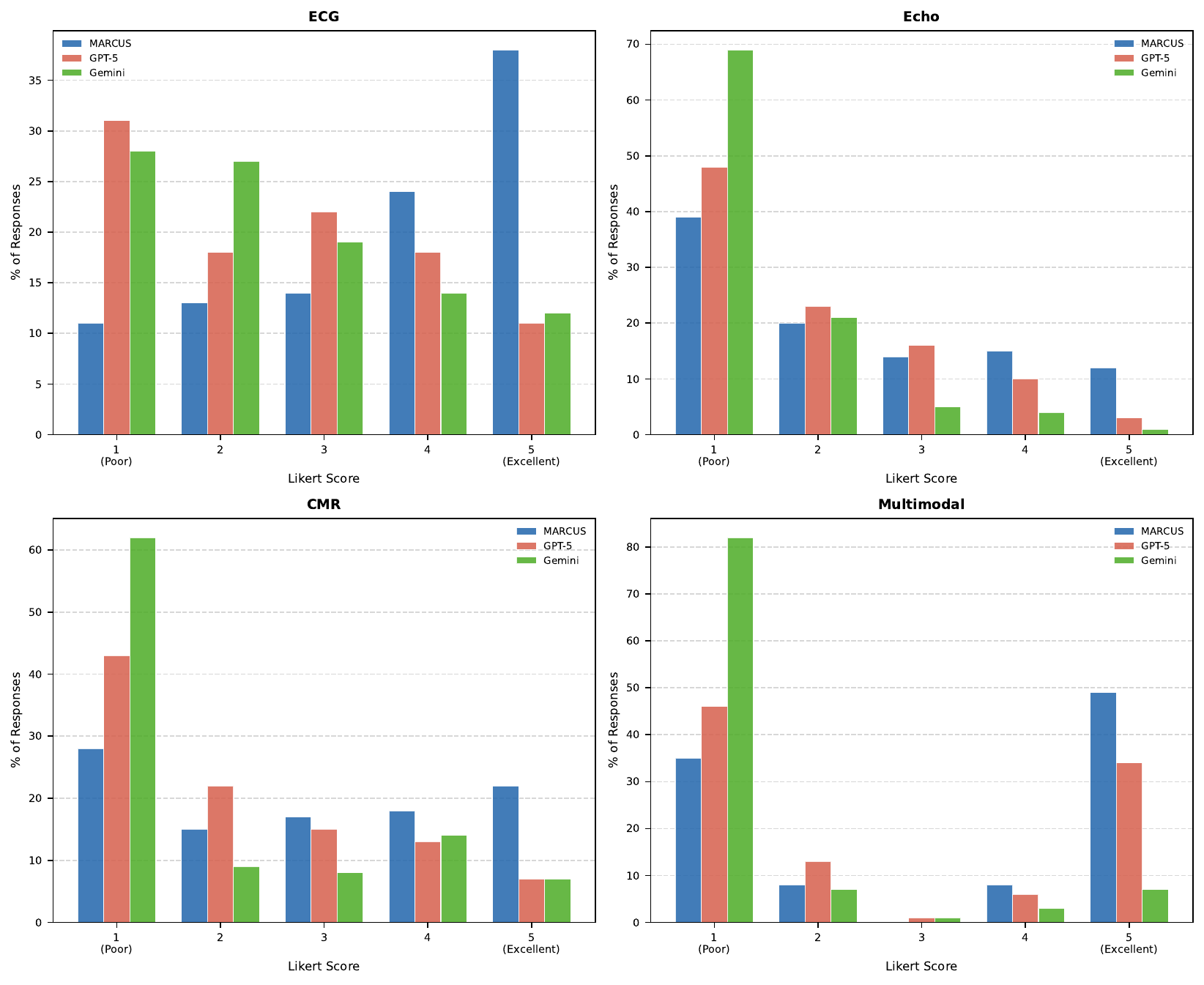}
\caption{\textbf{Frequency histograms of Likert score distributions.} Bars show the percentage of responses assigned each integer score (1--5) for MARCUS (blue), GPT-5 (red), and Gemini 2.5 Pro (green) within each modality. Scores 1--2 indicate clinically inadequate responses; 3 indicates acceptable; 4--5 indicate good to expert quality.}
\label{fig:suppS7}
\end{figure}

\section{MARCUS Failure Case Analysis: Error Type Distribution}

Across MCQ tasks, MARCUS returned 40 incorrect answers from 188 paired comparisons (overall failure rate 21.3\%), with modality-specific rates of 12.0\% for ECG (6/50), 36.0\% for echocardiography (18/50), 12.0\% for CMR (6/50), and 26.3\% for multimodal questions (10/38). The dominant MCQ error type was hallucination/fabrication ($n$=20, 50.0\% of MCQ failures).

For VQA tasks, 169 of the 400 MARCUS responses across modalities received a Likert score of $\leq$2, corresponding to a failure rate of 42.3\%. Failure rates by modality were: echocardiography 59.0\% (59/100), CMR 43.0\% (43/100), multimodal 43.0\% (43/100), and ECG 24.0\% (24/100). Modality confusion accounted for 23.1\% of VQA failures ($n$=39), visual misinterpretation for 20.1\% ($n$=34), and reasoning/synthesis errors for 11.8\% ($n$=20).

Taken together, these results highlight echocardiography as the modality where MARCUS has the greatest potential for improvement, particularly in free-text generation tasks.

\begin{figure}[H]
\centering
\includegraphics[width=\textwidth]{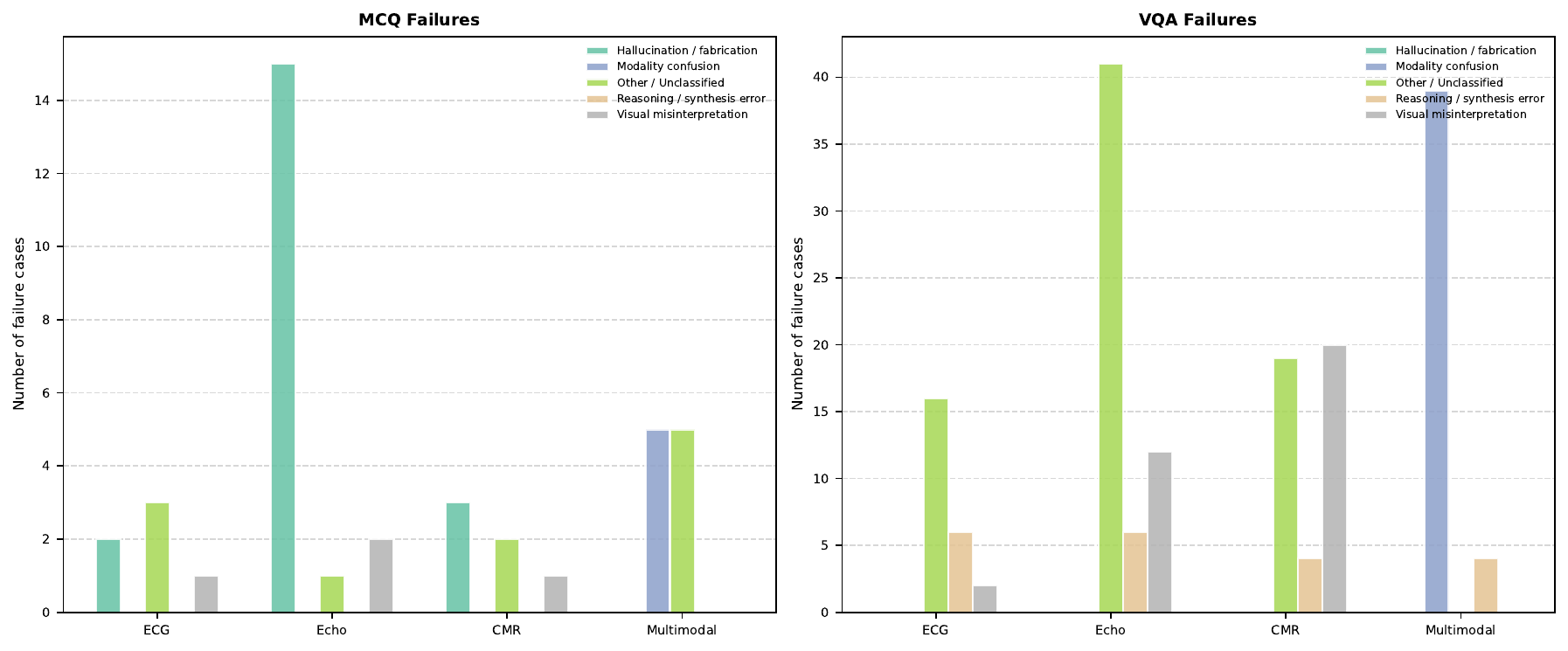}
\caption{\textbf{Error type distribution for MARCUS failure cases.} Left panel: MCQ failures ($n$=40 across all modalities), categorised by error type and stratified by modality. Right panel: VQA failures (Likert $\leq$2, $n$=169), with the same stratification.}
\label{fig:suppS8}
\end{figure}
\end{document}